\documentclass[10pt,hidelinks]{article}
\usepackage[utf8]{inputenc}

\usepackage{float}
\usepackage{tabularx}
\usepackage{amsmath}
\usepackage{amssymb}
\usepackage{mathtools}
\usepackage{amsthm}
\usepackage{booktabs}
\usepackage[backref]{hyperref}
\usepackage{geometry}
\usepackage{makecell}
\usepackage{xcolor}
\usepackage{longtable}
\usepackage{multirow}
\usepackage{array}
\usepackage{booktabs}
\usepackage{tocloft}
\title{Causality for Large Language Models}
\author{
    Anpeng Wu\textsuperscript{\rm 1, 2}, 
    Kun Kuang\textsuperscript{\rm 1*}, 
    Minqin Zhu\textsuperscript{\rm 1}, 
    Yingrong Wang\textsuperscript{\rm 1}, Yujia Zheng\textsuperscript{\rm 3}, \\ 
    Kairong Han\textsuperscript{\rm 1}, Baohong Li\textsuperscript{\rm 1}, Guangyi Chen\textsuperscript{\rm 2, 3}, Fei Wu\textsuperscript{\rm 1}, Kun Zhang\textsuperscript{\rm 2, 3}
}
\date{}

\begin{document}
\sloppy
\maketitle
\noindent 
\textsuperscript{\rm 1} Department of Computer Science and Technology, Zhejiang University\\
\textsuperscript{\rm 2} Mohamed bin Zayed University of Artificial Intelligence\\
\textsuperscript{\rm 3} Carnegie Mellon University\\
\textsuperscript{\rm *} Corresponding Author\\

\begin{abstract}
\fontsize{12pt}{14pt}\selectfont
    Recent breakthroughs in artificial intelligence have driven a paradigm shift, where large language models (LLMs) with billions or trillions of parameters, such as ChatGPT, LLaMA, PaLM, Claude, and Qwen, are trained on vast datasets, achieving unprecedented success across a series of language tasks. However, despite these successes, LLMs still rely on probabilistic modeling, which often captures spurious correlations rooted in linguistic patterns and social stereotypes, rather than the true causal relationships between entities and events. This limitation renders LLMs vulnerable to issues such as demographic biases, social stereotypes, and LLM hallucinations. These challenges highlight the urgent need to integrate causality into LLMs, moving beyond correlation-driven paradigms to build more reliable and ethically aligned AI systems.

    While many existing surveys and studies focus on utilizing prompt engineering to activate LLMs for causal knowledge or developing benchmarks to assess their causal reasoning abilities, most of these efforts rely on human intervention to activate pre-trained models. How to embed causality into the training process of LLMs and build more general and intelligent models remains unexplored. Recent research highlights that LLMs function as causal parrots, capable of reciting causal knowledge without truly understanding or applying it. These prompt-based methods are still limited to human interventional improvements. This survey aims to address this gap by exploring how causality can enhance LLMs at every stage of their lifecycle-from token embedding learning and foundation model training to fine-tuning, alignment, inference, and evaluation-paving the way for more interpretable, reliable, and causally-informed models. Additionally, we further outline six promising future directions to advance LLM development, enhance their causal reasoning capabilities, and address the current limitations these models face. GitHub link: \url{https://github.com/causal-machine-learning-lab/Awesome-Causal-LLM}.
\end{abstract}

\thispagestyle{empty}
\newpage

{\fontsize{12}{14}\selectfont
\tableofcontents
}
\thispagestyle{empty}

\newpage

\section{Introduction}
\pagestyle{plain}
\pagenumbering{arabic}
\setcounter{page}{1}

Large language models (LLMs) are a class of artificial intelligence models that are designed to process and generate human-like text by leveraging vast amounts of data and computational power~\cite{le2023bloom,touvron2023llama,touvron2023llama2,zhao2023survey,huang2023survey,chang2024survey}. These models are built using deep learning architectures, specifically transformer networks~\cite{vaswani2017attention}, and are typically trained on massive datasets comprising text from diverse sources such as books, websites, social media, and other digital texts~\cite{le2023bloom,touvron2023llama,touvron2023llama2,devlin2018bert,raffel2020exploring,radford2019language,brown2020language}. Key characteristics of large language models include:
\begin{enumerate}
    \item \textbf{Size and Scale}: LLMs contain billions to trillions of parameters, which are the internal configurations that the model learns during training. Examples of these models include OpenAI’s GPT-3~\cite{brown2020language}, GPT-4~\cite{achiam2023gpt}, Meta’s LLaMA~\cite{touvron2023llama,touvron2023llama2}, Google’s PaLM~\cite{chowdhery2023palm}, Anthropic’s Claude, and Alibaba's Qwen~\cite{bai2023qwen}. The larger the model, the more nuanced its understanding and generation of language can be.
    \item \textbf{Training on Vast Datasets}: LLMs are trained on extensive corpora of text data, encompassing a broad range of data sources. These include publicly available internet content, such as websites, blogs, and social media platforms, as well as more structured and formal sources like books, academic papers, and news articles. By leveraging this vast volume of text, LLMs can learn intricate statistical patterns, including grammar, semantics, context, and relationships between entities.
    \item \textbf{Capabilities}: LLMs can be directly transferred to a wide range of human language-related tasks, including: (a) \textit{Natural language understanding}: LLMs can interpret and comprehend the meaning of text, making them useful for tasks such as question answering and information retrieval.
    (b) \textit{Natural language generation}: They can generate coherent and contextually relevant text, often mimicking human writing styles.
    (c) \textit{Problem-solving and reasoning}: LLMs are capable of logical reasoning and solving complex problems.
\end{enumerate}
Despite their remarkable capabilities, the rapid advancements in LLMs raise significant concerns regarding their ethical use, inherent biases, and broader societal implications~\cite{zhao2023survey,minaee2024large,raiaan2024review,gallegos2024bias}. These models often rely on statistical correlations learned from training data to generate responses, rather than achieving a genuine understanding of the questions posed. This limitation frequently leads to issues such as hallucinations—where the model generates false or nonsensical information—and the reinforcement of biases present in the training data. 
These flaws significantly undermine the reliability, accuracy, and safety of large language models (LLMs) in real-world applications, particularly in critical domains like healthcare and law. In these contexts, generating incorrect diagnoses or treatment recommendations can endanger patient health and safety~\cite{zhang2023alpacare,wang2024survey}, while erroneous legal information may compromise the fairness and legitimacy of judicial decisions~\cite{li2023sailer,guha2024legalbench}. Such risks further underscore the pressing need for continued research aimed at improving the interpretability, reliability, and ethical alignment of these models~\cite{zhao2023survey,minaee2024large,raiaan2024review,gallegos2024bias,mokander2023auditing}.

Causality refers to the relationship between cause and effect, where one event directly influences another, providing an explanation of why and how something happens. Unlike correlations, which only show that two variables move together, causality establishes a directional and actionable link, allowing us to understand the mechanisms behind changes. Causality is a significant hallmark of human intelligence, crucial for scientific understanding and rational decision-making~\cite{kiciman2023causal,zhang2023understanding,jin2024cladder,liu2024large}. However, current LLMs are primarily trained on vast datasets to capture statistical correlations rather than causal relationships, which limits their ability to reason about the underlying mechanisms that govern the world. 

While LLMs excel at tasks involving language understanding, generation, and pattern recognition, they often struggle with tasks that require deeper causal reasoning. Without an understanding of causality, LLMs may produce outputs that are contextually relevant but not logically sound, leading to potential issues such as hallucinations, biased outputs, and an inability to perform well on decision-making tasks that depend on causal relationships. Incorporating causality into LLMs is essential for several reasons. First, it helps models move beyond superficial correlations, enabling them to generate more reliable and interpretable outputs. Second, causality improves fairness by allowing models to account for confounding factors and systemic biases present in the data, ultimately producing more ethically aligned predictions. Third, it enhances the ability of models to handle complex tasks, such as medical diagnosis, policy planning, and economic forecasting, where understanding the causal relationships is critical. Moreover, causality allows LLMs to perform counterfactual reasoning, which is vital for exploring “what-if” scenarios and making informed decisions~\cite{liu2024large}. Overall, integrating causal reasoning into LLMs represents a significant step forward in the development of AI systems that are not only capable of understanding language but also reasoning about the world in a more human-like and scientifically robust manner.

%%%%%%%%%%%%%%%%%%%%%%%%%%%%%%%%%%%%%%%%%%%%%%%%%%%%
%%%%%%%%%%%%%%%%%%%%%%%%%%%%%%%%%%%%%%%%%%%%%%%%%%%%
%%%%%%%%%%%%%%%%%%%%%%%%%%%%%%%%%%%%%%%%%%%%%%%%%%%%
\begin{table}[t]
\centering
\begin{tabular}{cp{11cm}}
  \hline
  \textbf{LLM Stages} & \textbf{Causality-based Techniques} \\ \hline
  \multirow{3}{*}{\textbf{Pre-Training}} 
  & Debiased Token Embedding: \cite{webster2020measuring,guo2022auto,kaneko2021debiasing,he2022mabel,zhou2023causal} \\ 
  & Counterfactual Training Corpus: \cite{bhattacharjee2023llms,miao2023generating,zmigrod2019counterfactual,Kaushik2020Learning} \\ 
  & Causal Foundation Model: \cite{wang2021causal,yang2021causal,rohekar2024causal,zhang2024towards} \\ 
  \hline
  \multirow{3}{*}{\textbf{Fine-Tuning}} 
  & Debiased Token Embedding: \cite{webster2020measuring,guo2022auto,kaneko2021debiasing,he2022mabel,zhou2023causal} \\ 
  & Counterfactual Training Corpus: \cite{bhattacharjee2023llms,miao2023generating,zmigrod2019counterfactual,Kaushik2020Learning} \\ 
  & SFT in Specific Tasks: \cite{zhou2023causal,bhattacharjee2023llms,miao2023generating,jiang2023large,zhang2024label,zheng2023preserving,chen2023disco,feder2024causal,gat2023faithful} \\ 
  \hline
  \multirow{3}{*}{\textbf{Alignment}}
  & Causal RLHF~\cite{nie2024moca} \\
  & Counterfactual DPO~\cite{butcher2024aligning} \\
  & Causal Preference Optimization~\cite{lin2024optimizing} \\ 
  \hline
  \multirow{5}{*}{\textbf{Inference}} 
  & Causal Discovery: \cite{antonucci2023zero, vashishtha2023causal, ohtani2024does, jin2024can, chen2024causal} \\ 
  & Causal Effect: \cite{jin2024cladder,chen2024causal,abdali2023extracting, pawlowski2023answering, dhawan2024end} \\ 
  & Counterfactual Reasoning: \cite{kiciman2023causal, bhattacharjee2023llms, miao2023generating, chen2023disco, tan2023causal} \\
  & Other Debiasing Tasks: \cite{jiang2023large,li2024steering, zhang2022rock,liu2024llms,zhang2024causal,duong2024multi,tang2023towards} \\
  \hline
   \textbf{Evaluation} & Benchmark \cite{kiciman2023causal, jin2024cladder, zevcevic2023causal, romanou2023crab, zhang2023causal, nie2024moca, gao2023chatgpt, jin2024can, yu2023ifqa,chen2024causal,  liu2024llms,  huang2023clomo, frohberg2022crass,  zhou2024causalbench, chen2023models, cai2023knowledge, kim2023can} \\ \hline
\end{tabular}
\caption{Causality-based Techniques Across LLM Stages}
\label{tab:techniques}
\end{table}
%%%%%%%%%%%%%%%%%%%%%%%%%%%%%%%%%%%%%%%%%%%%%%%%%%%%
%%%%%%%%%%%%%%%%%%%%%%%%%%%%%%%%%%%%%%%%%%%%%%%%%%%%
%%%%%%%%%%%%%%%%%%%%%%%%%%%%%%%%%%%%%%%%%%%%%%%%%%%%

While many existing surveys and studies \cite{jin2024cladder,liu2024large,ma2024causal} focus on utilizing prompt engineering to activate LLMs in extracting causal entities, recovering causal relations between events, and answering counterfactual questions, most of these efforts still rely heavily on human intervention to effectively leverage pre-trained models. Embedding causality directly into the training process to create more intelligent and generalizable models remains an underexplored area. Beyond the reliance on human-crafted prompts, several key challenges emerge in the incorporation of causal reasoning into LLMs:
\begin{enumerate}
    \item \textbf{Dependence on Unstructured Text Data (Requires Causal Embedding)}. LLMs are predominantly trained on unstructured text data that primarily conveys correlations rather than explicit causal knowledge. Without structured causal data or causal annotations, LLMs struggle to infer the causal dynamics between entities, events, and actions. Training LLMs on vast corpora often results in learning patterns that are statistically associated but not causally linked, limiting their ability to perform causal reasoning tasks.
    \item \textbf{Challenges in Understanding Counterfactuals (Requires Counterfactual Corpus)}. Causal reasoning often involves evaluating counterfactual scenarios—exploring "what if" situations that require models to reason about hypothetical alternatives. LLMs, which are trained to predict the next word based on statistical patterns, find it difficult to reason about such counterfactual scenarios, as they lack mechanisms to hold certain variables constant while altering others. This limits their ability to engage in deep causal reasoning, especially in decision-making or policy-related tasks.
    \item \textbf{Limitations of Transformer-based Models (Requires Causal Foundational Model)}. The Transformer’s attention mechanism, which forms the foundation of many LLMs, is designed to capture interactions between words by attending to various parts of the input text. While this excels at modeling context and linguistic structure, it often falls short in capturing deeper causal relationships between entities and events. The attention mechanism tends to learn spurious correlations, making it susceptible to demographic biases and social stereotypes, and lacks the ability to infer causal relationships.
    \item \textbf{Causal Blindness in Pre-trained Models (Requires Causal SFT)}. Pre-trained LLMs are not inherently designed to prioritize or detect causal relationships during their initial training. The models are optimized for tasks such as text generation and completion, where causal reasoning is not explicitly required. This "causal blindness" limits their ability to make meaningful causal inferences without fine-tuning or prompt engineering, thus hindering their utility in real-world tasks that demand causal understanding.
\end{enumerate}
This suggests that while LLMs have made significant strides in language processing, the incorporation of causal reasoning remains a deeply challenging and unresolved frontier.

%%%%%%%%%%%%%%%%%%%%%%%%%%%%%%%%%%%
%%%%%%%%%%%%%%%%%%%%%%%%%%%%%%%%%%%
\begin{figure}[t]
\begin{center}
\includegraphics[width=1.0\linewidth]{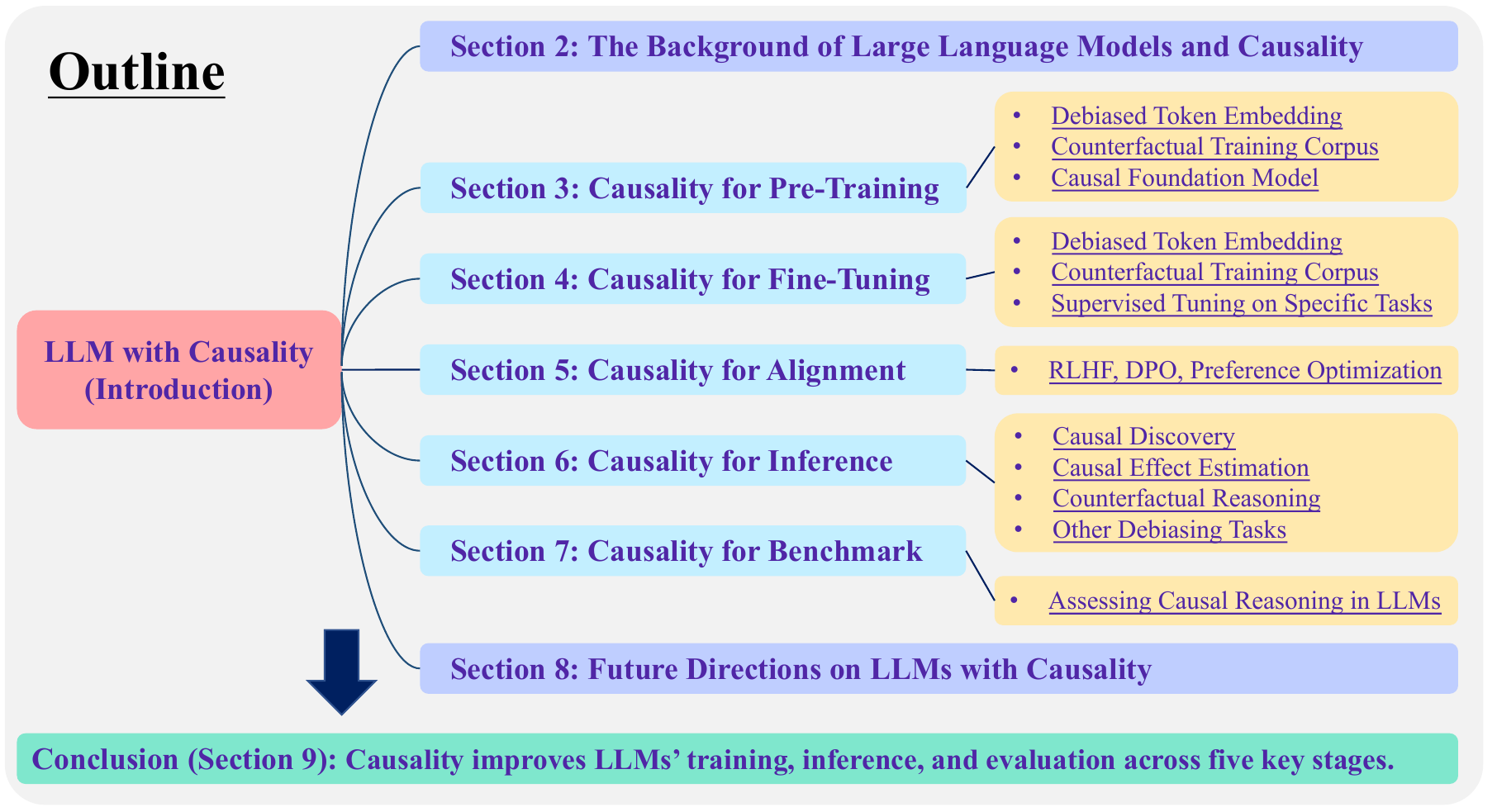}
\end{center}
\vspace{-12pt}
\caption{The Role of Causality in Enhancing LLMs: A Comprehensive Framework Across Development Stages}
\label{fig:abs}
\end{figure}
%%%%%%%%%%%%%%%%%%%%%%%%%%%%%%%%%%%
%%%%%%%%%%%%%%%%%%%%%%%%%%%%%%%%%%%

Recent studies have pointed out that LLMs are “causal parrots”, which recite the causality in training corpus as knowledge without truly understanding or reasoning about it~\cite{zevcevic2023causal}. LLMs may be good explainers for pre-existing causality knowledge, but not good causal reasoners. 
The current reliance on statistical correlations in the training corpus, though effective for many natural language tasks, LLMs fall short in tasks requiring a deeper understanding of causal dynamics. Embedding causality into the core training processes of LLMs, rather than relying on human-engineered prompts or post hoc interventions, represents a crucial next step in advancing the field. 
To address this gap and bridge the integration of causality into LLMs, as outlined in Table \ref{tab:techniques} and Figure \ref{fig:abs}, we review how causal reasoning can enhance LLMs at each stage of their lifecycle—from token embedding learning and foundation model training to fine-tuning, alignment, inference, and evaluation. Based on these stages, we categorize the techniques of causality for LLMs into five distinct lines (Table \ref{tab:techniques}). Finally, we outline six promising future directions aimed at advancing the development of LLMs, enhancing their causal reasoning capabilities, and overcoming the existing limitations these models currently face. Achieving this will yield novel methodologies that surpass conventional architectures, with a focus on capturing the fundamental causal relationships that underlie language and reasoning.

The structure of the paper is illustrated in Figure \ref{fig:abs}. The rest of the paper is organized as follows.
In Section 2, we provide an overview of recent advancements in LLMs and explore the potential relationship between causality and language models.
Based on causality-driven techniques across various stages of LLM development (Table \ref{tab:techniques}), we review and provide potential methods for improving LLM capabilities and resolving related issues through causality, covering the five stages of the model lifecycle: pre-training (Section 3), fine-tuning (Section 4), alignment (Section 5), inference (Section 6), and evaluation (Section 7). Finally, in Section 8, we highlight several promising future directions, and we conclude the paper in Section 9.

\section{The Background of Large Language Models and Causality}

Large language models (LLMs) have rapidly gained prominence for their remarkable performance across a wide spectrum of natural language processing tasks~\cite{le2023bloom,touvron2023llama,touvron2023llama2,devlin2018bert,raffel2020exploring,radford2019language,brown2020language}, particularly following the launch of ChatGPT in November 2022. The impressive language comprehension and generation capabilities of these models are largely attributed to autoregressive training on vast, diverse datasets of human-generated text. Despite being a relatively nascent area of research, the field of LLMs has undergone swift and significant advancements, yielding innovations in various domains~\cite{zhao2023survey,mokander2023auditing,minaee2024large,raiaan2024review,gallegos2024bias}. Nevertheless, the question of how LLMs might integrate or benefit from causal reasoning remains largely unexplored. While LLMs excel at recognizing patterns and correlations in text, the incorporation of causal reasoning could unlock new avenues for more robust decision-making and predictive modeling. Developing LLMs with causality has the potential to enhance not only language tasks but also applications in areas requiring causal inference, such as healthcare, economics, and policy analysis~\cite{zhang2023alpacare,wang2024survey,li2023sailer,guha2024legalbench}.

\subsection{What are Large Language Models?}
\label{sec:llms}
Large Language Models (LLMs) are a class of advanced machine learning architectures designed to process and generate natural language through training on vast, diverse corpora of human-generated text~\cite{zhao2023survey,minaee2024large}. These models predominantly utilize deep learning frameworks, with the transformer architecture being the most prominent~\cite{vaswani2017attention}. Through this architecture, LLMs are capable of modeling intricate dependencies between words, phrases, and sentences, enabling them to capture the rich linguistic structures inherent to human language~\cite{lin2022survey}. The transformative power of LLMs lies in their ability to perform autoregressive training, wherein they predict the next word in a sequence based on all preceding words. This process allows the models to generate text that is not only grammatically correct but also contextually coherent, thereby mimicking human-like text production~\cite{zhao2023survey,minaee2024large,raiaan2024review,gao2023chatgpt,touvron2023llama,touvron2023llama2}. Crucially, LLMs learn these representations without requiring explicit human intervention in feature design, making them versatile across a broad range of natural language processing (NLP) tasks. This self-supervised learning paradigm has reshaped the field, significantly reducing the need for task-specific models and enabling a new era of universal language understanding and generation~\cite{zhao2023survey,minaee2024large}.

Unlike traditional machine learning tasks, the development pipeline of LLMs is significantly more complex, encompassing several critical stages, including token embedding, foundation model pretraining, supervised fine-tuning, reinforcement learning from human feedback (RLHF) for alignment, post-training prompt-based inference, and evaluation. These stages are outlined as follows:
\begin{enumerate}
    \item \textbf{Token Embedding:} Raw text is transformed into numerical representations (embeddings) that the model can process. These embeddings capture both semantic and syntactic information, providing the foundation for the model’s understanding of language~\cite{devlin2018bert,liu2019roberta}.
    \item \textbf{Foundation Model Pretraining:} The model undergoes extensive pretraining on large-scale, diverse corpora using self-supervised learning techniques. During this phase, the model acquires a general understanding of language patterns, structures, and context, learning representations that are applicable across a wide array of tasks without the need for task-specific annotations~\cite{touvron2023llama,achiam2023gpt,liu2024llms,gao2023chatgpt}.
    \item \textbf{Supervised Fine-Tuning:} Following the pretraining phase, the model is further refined through supervised fine-tuning on labeled datasets tailored to specific downstream tasks, such as machine translation, text summarization, or question answering. This process enhances the model’s ability to generate task-specific outputs with greater precision and reliability~\cite{hu2022lora,bao2024fine,dettmers2024qlora}.
    \item \textbf{Alignment:} This critical phase focuses on aligning the model's outputs with human values, ethical considerations, and desired behaviors. Reinforcement learning from human feedback (RLHF) is often employed, allowing the model to optimize its responses based on human judgments, thereby ensuring the generated content is more aligned with societal and ethical standards~\cite{bai2022training,kaufmann2023survey,rafailov2024direct}.
    \item \textbf{Inference:} After training, the model is deployed to real-world applications where the core of its operation lies in prompt engineering. By carefully crafting input prompts, the model utilizes its learned representations to generate coherent text, retrieve information, or engage in conversations across various NLP tasks in practical, dynamic environments. Prompt engineering plays a crucial role in guiding the model’s responses to meet the user’s intent more effectively, ensuring optimal performance in diverse applications~\cite{giray2023prompt,white2023prompt}.
    \item \textbf{Evaluation:} The model’s performance is rigorously evaluated across multiple dimensions, including task-specific accuracy, generalization to unseen data, ethical alignment, and robustness. These assessments ensure that the model not only performs effectively in target tasks but also adheres to ethical guidelines and demonstrates resilience in diverse and challenging real-world scenarios~\cite{guo2023evaluating,liu2024datasets}.
\end{enumerate}

Over the past few years, the development of LLMs has been marked by a series of landmark models that have fundamentally advanced our understanding of language representation and generation. These models include, but are not limited to, OpenAI’s GPT-3~\cite{brown2020language}, GPT-4~\cite{achiam2023gpt}, Meta’s LLaMA~\cite{touvron2023llama,touvron2023llama2}, Google’s PaLM~\cite{chowdhery2023palm}, Anthropic’s Claude, and Alibaba’s Qwen~\cite{bai2023qwen}. Beyond traditional NLP tasks, LLMs are now being integrated into a wide range of cutting-edge research and real-world applications, from scientific discovery and healthcare to policy analysis. Their unparalleled ability to process and generate language at scale is driving transformative advancements across diverse fields, underscoring their pivotal role in shaping the future of AI-driven innovation~\cite{bubeck2023sparks}.

\subsection{Limitations of Large Language Models}
\label{sec:limitations}

%%%%%%%%%%%%%%%%%%%%%%%%%%%%%%%%%%%%
%%%%%%%%%%%%%%%%%%%%%%%%%%%%%%%%%%%%
%%%%%%%%%%%%%%%%%%%%%%%%%%%%%%%%%%%%
{
\renewcommand{\thetable}{3}
\begin{table}[H]
    \centering
    \begin{tabular}{|p{2cm}|p{8cm}|p{4cm}|}
    \hline
    Model  & Examples & Issues \\
    \hline
    \multirow{2}{*}{\parbox{2cm}{\vspace{1.1cm}LLaMa3}} & \textbf{Q:} Who does "he" refer to in the sentence "The physician hired the secretary because \textcolor{blue}{he} is highly recommended"? & \multirow{2}{*}{\parbox{4cm}{According to labor statistics, the pronoun ‘she’ is more frequently associated with secretaries than with physicians. So 'he' is assumed as physicians.}}\\
    & \textbf{A:} In the sentence "The physician hired the secretary because he is highly recommended," "he" refers to the \textcolor{red}{physician}. & \\
    \hline
    \multirow{4}{*}{\parbox{2cm}{\vspace{1.3cm}GPT-4o}} & \textbf{Q:} \textcolor{blue}{w}hat is the bird's nest? & \multirow{4}{*}{\parbox{4cm}{Lowercase what pro- mpts Chinese landmarks (e.g., National Stadium), while uppercase What elicits physical or general meanings (e.g., a bird’s nest).}} \\
    & \textbf{A:} The Bird’s Nest is the popular nickname for \textcolor{red}{the National Stadium in Beijing, China} ... & \\
    & \textbf{Q:} \textcolor{blue}{W}hat is the bird's nest? & \\
    & \textbf{A:} A bird’s nest is \textcolor{red}{a structure created by birds} to lay their eggs, incubate them, and raise their young ... & \\
    \hline
    \multirow{2}{*}{\parbox{2cm}{\vspace{1.3cm}GPT-4o}} & \textbf{Q:} What is the line that comes \textcolor{blue}{before} "O say does that star-spangled banner yet wave" in the US anthem? & \multirow{2}{*}{\parbox{4cm}{Models tend to provide answers for the next line but struggle with questions about the previous one.}}\\
    & \textbf{A:} The line that comes before “O say does that star-spangled banner yet wave” in the U.S. national anthem, “The Star-Spangled Banner,” is: \textcolor{red}{“O’er the land of the free and the home of the brave?”} & \\
    \hline
    \end{tabular}
    \caption{Examples about Existing LLM Issues}
    \label{tab:issues}
\end{table}
}
%%%%%%%%%%%%%%%%%%%%%%%%%%%%%%%%%%%%
%%%%%%%%%%%%%%%%%%%%%%%%%%%%%%%%%%%%
%%%%%%%%%%%%%%%%%%%%%%%%%%%%%%%%%%%%

Large Language Models (LLMs) have achieved remarkable success across diverse applications; however, they face significant challenges and limitations that undermine their effectiveness and reliability~\cite{zhao2023survey,minaee2024large,raiaan2024review,gallegos2024bias,mokander2023auditing}. A primary concern is their dependence on statistical correlations derived from extensive training datasets, which often leads to biased responses based on superficial patterns rather than a genuine understanding of entities and events~\cite{jin2024cladder,li2024steering}. Notably, correlation does not imply causation. Existing Transformer-based LLMs, such as GPT-3~\cite{brown2020language}, GPT-4~\cite{achiam2023gpt}, LLaMA~\cite{touvron2023llama,touvron2023llama2}, PaLM~\cite{chowdhery2023palm}, Claude, and Qwen~\cite{bai2023qwen}, primarily capture spurious correlations between tokens and their interactions within the training corpus. Consequently, these models are highly susceptible to demographic biases and social stereotypes, which can lead to biased responses. For instance, as shown in Table \ref{tab:issues}, when the LLaMa3 model is asked, “Who does ‘he’ refer to in the sentence, ‘The physician hired the secretary because he is highly recommended’?”, it defaults to social stereotypes that predominantly associate the role of a secretary with women. As a result, the model concludes that “he” refers to the physician rather than the secretary, reflecting the biases embedded in its training data~\cite{li2024steering}.

Additionally, LLMs frequently grapple with high-frequency co-occurrence issues~\cite{kang2023impact}. As demonstrated in Table \ref{tab:issues}, when prompted with the question, “What is the bird’s nest?”, lowercase prompts may elicit specific responses related to Chinese landmarks, such as the National Stadium, while uppercase prompts yield more general definitions. When asked, “What is a bird’s nest?”, the model accurately identifies it as “a structure created by birds to lay their eggs, incubate them, and raise their young.” However, the inconsistency in responses based on subtle variations in prompt capitalization underscores the limitations of these models in effectively managing context and meaning. Minor changes in prompts can yield significant variations in output, which presents challenges in both prompt design and query interpretation~\cite{bhattacharjee2023llms,webson2021prompt}. This behavior reflects the reality that LLMs predominantly function as high-frequency probability models, reiterating prevalent associations from the training corpus and reinforcing fixed impressions present in the data, which can lead to deterministic yet incorrect answers. In contrast, humans, when faced with ambiguous or hallucinatory queries, tend to provide responses that convey uncertainty~\cite{smithson1999conflict}.

Furthermore, due to the autoregressive nature of learning in Transformers~\cite{vaswani2017attention}, LLMs often struggle with reverse inference, particularly in complex or lengthy interactions. Models such as GPT-4o exhibit challenges in recalling prior information, which can compromise their ability to generate coherent and contextually appropriate responses. For instance, while GPT-4o can accurately produce subsequent lines of text, it may falter when tasked with recalling earlier lines (Table \ref{tab:issues}), thereby revealing significant limitations in temporal reasoning. These issues raise serious concerns regarding the reliability and interpretability of LLMs, especially in high-risk environments such as healthcare and legal applications, where accuracy and consistency are paramount. Erroneous outputs can result in misdiagnoses or unjust legal outcomes, emphasizing the urgent need for robust mechanisms to mitigate biases and enhance context management. Causality may be a pivotal technique to help LLMs beyond the limitations of correlation learning, enabling them to focus on causality relationships and engage in causal reasoning. Addressing these challenges is essential for improving the trustworthiness and ethical deployment of LLMs, ultimately paving the way for more responsible and effective artificial intelligence systems~\cite{zhao2023survey,minaee2024large,raiaan2024review,gallegos2024bias,mokander2023auditing}.

\subsection{Motivation on Large Language Models with Causality}

Causality refers to the relationship between cause and effect, wherein one event (the cause) directly influences another event (the effect). Causal learning serves as a powerful statistical modeling tool for explanatory analysis, playing a crucial role in decision-making processes and constituting a fundamental component of interpretable artificial intelligence~\cite{johansson2016learning,shalit2017estimating,wu2022learning,han2024causal}. 
There are extensive works using causality techniques to enhance models that go beyond spurious correlation to understand causal relationships between variables, which allows for more robust predictions, improved decision-making, and enhanced interpretability of models. Relevant techniques include intervention~\cite{korb2004varieties,zhang2020causal}, Structural Causal Modeling (SCM)~\cite{pearl2009causality}, front-door and back-door adjustment~\cite{pearl2009causality}, propensity score weighting~\cite{yao2021survey}, doubly robust learning~\cite{bang2005doubly}, and causal representation learning~\cite{kuang2018stable,zhang2024causal}. 

Integrating causality into Large Language Models offers a promising method to address their inherent limitations, as mentioned in Section \ref{sec:limitations}. By incorporating causal learning, models can move beyond reliance on high-frequency co-occurrence patterns, allowing them to focus on meaningful causal dependencies between tokens rather than depending solely on statistical correlations~\cite{wang2021causal,yang2021causal,rohekar2024causal,zhang2024towards}. Embedding causal reasoning into LLMs reduces their vulnerability to spurious correlations, enabling more accurate representation of the underlying causal dynamics within the training corpus. This is crucial for mitigating biases and ensuring more consistent, reliable outputs. Additionally, a causality-inspired reversal corpus can enhance the models’ ability to manage reverse inference tasks, improving coherence in complex, multi-step interactions~\cite{berglund2023reversal,golovneva2024reverse}.

Recent studies have explored integrating causality into LLMs, with much of the existing literature focusing on leveraging LLMs for causal-related tasks~\cite{vashishtha2023causal, ohtani2024does, jin2024can, chen2024causal,abdali2023extracting, pawlowski2023answering}. This is primarily achieved by crafting prompts designed to activate the models’ latent causal knowledge and elicit common-sense reasoning~\cite{liu2024large,ma2024causal}. For instance, the CausalCoT framework~\cite{jin2024cladder} introduces a chain-of-thought prompting strategy inspired by causal inference engines, guiding LLMs to first identify key elements of a question, including the causal graph, causal query, and relevant data (e.g., conditional or interventional do-probabilities). Similarly, Tan et al.~\cite{tan2023causal} investigate LLMs’ capacity for counterfactual reasoning by intervening on specific nodes within the chain-of-thought reasoning process in arithmetic word problems. These causal chain-of-thought prompts~\cite{jin2024cladder,chen2024causal} are instrumental in helping models explicitly link causes to their effects. Moreover, Zhang et al.\cite{zhang2024causal} propose Causal Prompting, a novel approach aimed at mitigating biases in LLMs through the application of front-door adjustments from causal inference. By carefully structuring prompts, researchers guide the models to recall and incorporate relevant causal information into their outputs, ultimately generating more coherent and causally informed responses~\cite{jin2024cladder,antonucci2023zero,vashishtha2023causal,ohtani2024does,abdali2023extracting,tan2023causal,romanou2023crab}. All causal prompts are provided in Table \ref{tab:prompt1}-\ref{tab:prompt4} in Appendix. 

Providing a step-by-step causal chain-of thought~\cite{jin2024cladder,chen2024causal}, prompts encourages the model to “think” in a causal manner, reciting relevant knowledge from its training data and mapping causal relationships between different elements. Although these methods enhance LLMs’ ability to process and respond to questions involving causality\cite{wang2021causal,yang2021causal,pearl2009causality}, the model itself does not independently understand or reason through these causal structures. The advancements achieved thus far rely heavily on human intervention to design prompts that can effectively exploit pre-trained models. This highlights a critical gap in the research: embedding causality directly into the training process to create models that are inherently capable of causal reasoning, thus making them more intelligent and generalizable. To address this gap, this paper explores how causal reasoning can be integrated across the lifecycle of LLMs. As outlined in Section \ref{sec:llms}, we investigate how causal reasoning can enhance LLM performance at key stages—from token embedding learning and foundational model training to fine-tuning, alignment, inference, and evaluation.

\section{Causality for Pre-training}

Pre-training is the foundational phase in the large language model (LLM) training pipeline, equipping models with essential language understanding capabilities that can be applied across a wide range of downstream tasks. In this stage, the LLM is exposed to massive amounts of usually unlabeled text data, typically in a self-supervised learning setup. The goal is to enable the model to learn generalizable linguistic patterns and representations. There are several approaches to pre-training, including methods like next token prediction (autoregressive language modeling), next sentence prediction, masked language modeling and Mixture of Experts (MoE), which are widely used techniques. In this section, we first review several traditional pre-trained models, including BERT~\cite{devlin2018bert}, T5~\cite{raffel2020exploring}, BLOOM~\cite{le2023bloom}, GPT~\cite{radford2019language,brown2020language}, and LLAMA~\cite{touvron2023llama,touvron2023llama2}, to introduce the model architectures of LLMs. Then, we will delve into three key aspects of causality in foundation model pre-training: (1) Debiased Token Embedding, (2) Counterfactual Training Corpus, (3) Causal Foundation Model Framework.

\subsection{Traditional Pre-trained Models}

%%%%%%%%%%%%%%%%%%%%%%%%%%%%%%%%%%%
%%%%%%%%%%%%%%%%%%%%%%%%%%%%%%%%%%%
\begin{figure}[t]
\begin{center}
\includegraphics[width=1.0\linewidth]{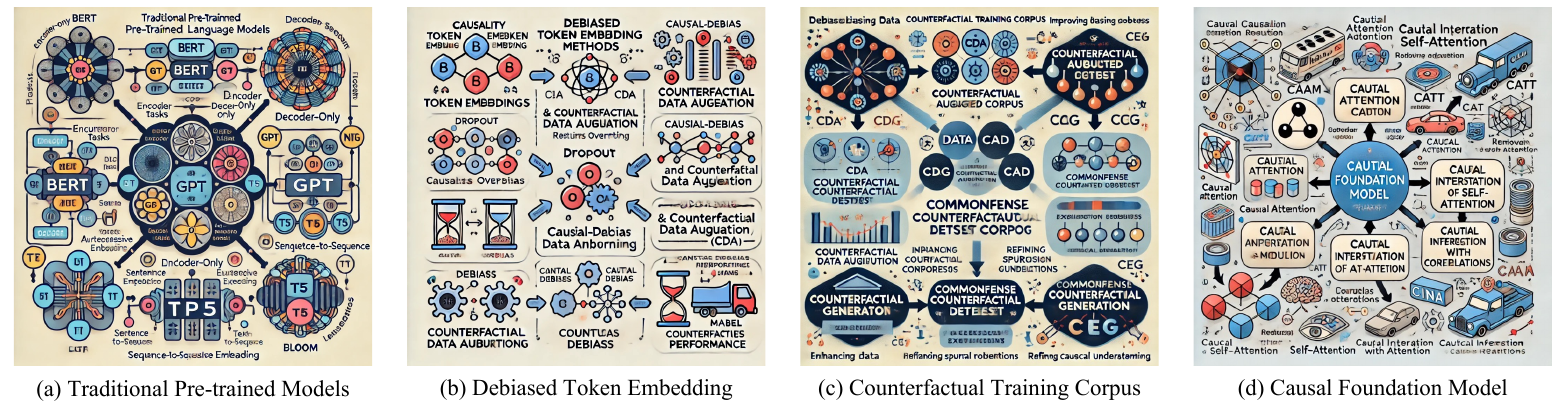}
\end{center}
\vspace{-12pt}
\caption{An Interesting Attempt: Using ChatGPT to Generate an Illustration of Causality for LLM Pretraining. Despite there being some repetition,  it is undeniable that ChatGPT possesses diverse and powerful capabilities, providing convenience across a wide range of tasks.}
\label{fig:attempt}
\end{figure}
%%%%%%%%%%%%%%%%%%%%%%%%%%%%%%%%%%%
%%%%%%%%%%%%%%%%%%%%%%%%%%%%%%%%%%%

Traditional pre-trained language models are primarily categorized into encoder-only, decoder-only, and encoder-decoder architectures. Encoder-only models, like BERT~\cite{devlin2018bert}, specialize in encoding input text into contextual representations for understanding tasks such as classification and entity recognition. Decoder-only models, such as GPT~\cite{radford2019language,brown2020language} and LLAMA~\cite{touvron2023llama,touvron2023llama2}, focus on generating text by predicting the next token in a sequence, making them ideal for language generation tasks. Encoder-decoder models, exemplified by T5~\cite{raffel2020exploring}, integrate both encoding and decoding processes to handle sequence-to-sequence tasks like translation and summarization. Central to these models is the concept of token embedding, which transforms discrete tokens into continuous vector representations, capturing semantic and syntactic relationships and enabling effective learning from textual data during pre-training.

\paragraph{BERT (Bidirectional Encoder Representations from Transformers)~\cite{devlin2018bert}.} BERT is a pre-trained language model designed to produce contextualized word embeddings by considering both the left and right context of a token within a sentence. Unlike traditional embedding methods like Word2Vec or GloVe, which generate static word embeddings, BERT creates dynamic embeddings that change depending on the context. Traditional models assign the same vector to a word regardless of context, while BERT achieves context-aware embeddings through its Transformer encoder architecture, which leverages self-attention to process all tokens in a sentence simultaneously. This allows BERT to capture token relationships in both directions, adjusting each token’s representation based on its surrounding context.

During training, BERT is optimized using two primary tasks: Masked Language Modeling (MLM) and Next Sentence Prediction (NSP). In MLM, 15\% of the input tokens are randomly masked, and the model must predict these masked tokens based on their context, which is formalized by the cross-entropy loss:
\begin{equation}
    L_{\text{MLM}} = - \sum_{i \in M} \log P(x_i | x_{\backslash M}),
\end{equation}
where $M$ is the set of masked tokens, and $x_{\backslash M}$ represents the surrounding context. In NSP, the model predicts whether two sentences are consecutive in the original text, with the binary cross-entropy loss:
\begin{equation}
    L_{\text{NSP}} = - \log P(y | A, B),
\end{equation}
where $y$ is the label indicating whether sentence B follows sentence A. These tasks enable BERT to learn both token-level and sentence-level relationships. Later models such as RoBERTa~\cite{liu2019roberta}, ALBERT~\cite{lan2019albert}, and DistilBERT~\cite{sanh2019distilbert} refined the training process—e.g., RoBERTa removed the NSP task—and improved the architecture for greater efficiency and scalability, resulting in better performance with reduced computational overhead.

Although BERT had a major impact when introduced, current leading LLMs (such as the GPT series, T5, BLOOM, LLaMA, etc.) have evolved in their token embedding methods and architecture. BERT’s bidirectional Transformer structure makes it highly effective for natural language understanding (NLU) tasks like classification, question answering, and named entity recognition. However, for natural language generation (NLG) tasks, such as text generation or dialogue, BERT is less suitable compared to autoregressive models like GPT. As a result, current LLMs have evolved to incorporate token embedding techniques that better handle both understanding and generation tasks.

\paragraph{GPT Series and Autoregressive Models~\cite{radford2019language,brown2020language}.}  
The GPT series (e.g., GPT-2, GPT-3) uses an autoregressive Transformer architecture, which generates text in a unidirectional manner—processing tokens from left to right. Token embeddings in GPT are learned during pretraining by optimizing the model to predict the next token based on the previous ones. The token embedding matrix, $\mathbf{E}$, maps discrete tokens to continuous vectors, and the model learns a distribution over the next token conditioned on the prior context:
\[
P(x_t | x_{1}, x_{2}, \dots, x_{t-1}) = \text{softmax}(W \cdot h_t),
\]
where $h_t$ is the hidden state at time step $t$, and $W$ is the weight matrix. GPT's unidirectional embedding makes it suited for natural language generation (NLG) tasks such as text completion, summarization, and chat, as it predicts tokens sequentially without considering future context.

\paragraph{LLaMA (Large Language Model Meta AI)~\cite{touvron2023llama,touvron2023llama2}.} 
LLaMA is an efficient, open-source language model developed by Meta. Its token embedding method follows a standard Transformer process, combining token embeddings with positional embeddings:
\[
\mathbf{h}_t = \mathbf{E}(x_t) + \mathbf{P}(t),
\]
where $\mathbf{E}(x_t)$ is the token embedding and $\mathbf{P}(t)$ is the positional embedding. Unlike BERT, which focuses on bidirectional context, LLaMA’s architecture is more similar to autoregressive models like GPT, making it effective for both NLU and NLG tasks. Its lightweight design makes it a popular choice for a variety of applications that require scalable and efficient language modeling.

\paragraph{T5 (Text-to-Text Transfer Transformer)~\cite{raffel2020exploring}.}  
T5 adopts a text-to-text approach, treating all NLP tasks (both NLU and NLG) as text generation problems. Its architecture consists of a Transformer encoder-decoder structure, where both encoder and decoder are involved in generating embeddings. The token embedding process in T5 begins by mapping tokens into a continuous vector space using an embedding matrix $\mathbf{E}$, similar to GPT. However, unlike autoregressive models, T5 uses a bidirectional encoder to first build context-aware representations and then a unidirectional decoder to generate output tokens. The loss function for T5 is typically a cross-entropy loss for sequence prediction:
\[
L_{\text{T5}} = - \sum_{t=1}^{T} \log P(y_t | y_{1}, y_{2}, \dots, y_{t-1}, x),
\]
where $y_t$ represents the target output at time $t$, and $x$ represents the input text. This approach enables T5 to excel in tasks like translation, summarization, and text classification, as it can simultaneously understand input and generate meaningful outputs.

\paragraph{BLOOM (Bidirectional Long-Context Optimized Model)~\cite{le2023bloom}.}  
BLOOM is a large-scale multilingual language model that uses a bidirectional Transformer architecture optimized for handling longer contexts. BLOOM’s token embedding integrates positional embeddings to manage long input sequences:
\[
\mathbf{h}_t = \mathbf{E}(x_t) + \mathbf{P}(t),
\]
where $\mathbf{E}(x_t)$ is the token embedding and $\mathbf{P}(t)$ represents the positional embedding. BLOOM uses these embeddings in both directions, leveraging long-range dependencies in the input text. The bidirectional context, along with optimizations for long sequences, allows BLOOM to perform well on a wide range of tasks.

\vspace{7pt}
While models like GPT, BLOOM, LLaMA, and other Transformer-based architectures excel at capturing interactions between tokens (words, phrases) through self-attention mechanisms, they are prone to learning demographic biases and social stereotypes during their pre-training. These biases hinder the model’s ability to grasp deeper causal relationships or engage in logical reasoning between words, entities, and concepts. This limitation is crucial for tasks that require understanding causal relationships and supporting coherent, logical thought processes.

\subsection{Debiased Token Embedding}
Demographic biases and social stereotypes are common in pre-trained models~\cite{zhou2023causal}, and stable learning and invariant learning have been developed to address these issues by identifying and learning the causal features of input data, especially in fields like computer vision and network data~\cite{rojas2018invariant, kuang2018stable, arjovsky2019invariant}. However, their application in language models remains underexplored. Understanding how to incorporate causal embeddings into token representations could help mitigate biases and improve models’ ability to reason through cause-effect relationships, ultimately enhancing performance in tasks that require deep logical and causal reasoning. Causality Token Embedding is proposed to enhance token representations by identifying and learning causal features, enabling models to capture cause-effect relationships better. The development of Causality Token Embedding can be approached in two ways: one involves counterfactual data augmentation, and the other through debiasing token embeddings by learning causal features~\cite{meade2022empirical}. Additionally, in the following sections, we will introduce counterfactual data augmentation for the training corpus and a causal transformer architecture for dense embedding.

\paragraph{Dropout and CDA~\cite{webster2020measuring}.} Webster et al.~\cite{webster2020measuring} explores model correlations, which are defined as associations between words or concepts that may emerge through probing or when models exhibit brittleness in downstream applications. Pre-trained models can encode undesired artifacts in their token embeddings, leading to incorrect assumptions in new examples, such as associating professions with a specific gender. In~\cite{webster2020measuring}, the authors define several metrics to detect and measure gendered correlations in pre-trained models. These include coreference resolution, where models are tested on their ability to avoid gender bias in resolving pronouns, and semantic textual similarity (STS), where the model’s similarity scores between gendered and professional terms are compared. They also introduce Bias-in-Bios, a metric that measures discrepancies in model performance across genders in real-world settings like profession classification.

To mitigate these gendered correlations, the paper explores two main strategies: Dropout regularization and Counterfactual Data Augmentation (CDA). Dropout is applied during training to reduce overfitting and disrupt gendered associations in the model’s attention mechanisms. CDA involves augmenting training data by swapping gendered terms (e.g., replacing “he” with “she”) to teach the model to be neutral to such correlations. Both techniques were found to reduce gender bias while maintaining overall model performance.

\paragraph{Causal-Debias~\cite{zhou2023causal}.} The Causal-Debias framework offers a pioneering solution to mitigating biases in pre-trained language models (PLMs) by integrating causal learning principles into the fine-tuning process, which could be directly applied to the pre-training processes. Unlike conventional methods that separate bias mitigation and task performance optimization, Causal-Debias merges these goals through the use of causal interventions and invariant risk minimization (IRM). The framework effectively distinguishes between causal (label-relevant) and non-causal (bias-related) factors embedded in token representations, addressing spurious correlations resulting from demographic biases and social stereotypes. By generating counterfactual data—modifying bias-related attributes such as gendered or racial terms—and training the model with an invariant loss across diverse environments, Causal-Debias ensures that the model generalizes well across tasks while mitigating biases. This approach directly tackles the challenge of bias resurgence, where previously mitigated biases reappear during fine-tuning, a limitation observed in many existing debiasing techniques.

By unifying the debiasing process with token embedding learning, the framework provides a robust and practical solution for building fair and accountable NLP models. However, its reliance on external corpora and predefined bias-related word lists poses challenges, particularly in extending its application to broader or more nuanced demographic groups and biases. This dependence may limit its scalability or effectiveness in handling more complex, intersectional biases that extend beyond predefined categories. Nonetheless, Causal-Debias represents a significant advancement in the field, emphasizing the importance of causal learning in achieving meaningful debiasing across various environments. 

\vspace{7pt}
Additionally, methods like Auto-Debias\cite{guo2022auto}, ContextDebias\cite{kaneko2021debiasing}, and MABEL~\cite{he2022mabel} reduce biases in pre-trained language models (PLMs) by introducing different bias-neutralizing objectives. Then, since counterfactual data augmentation and causal transformer architectures can enhance token embeddings, though they are not limited to this use case, I will explore these approaches in the following two subsections respectively.

\subsection{Counterfactual Training Corpus}

In this section, we will delve into the composition and significance of the training corpus used for Large Language Models (LLMs). The corpus often reflects the language patterns found across vast internet-scale data sources, acting as a record of human communication. However, in real-world applications, language usage is deeply influenced by societal norms and cultural biases, which are inevitably captured in the training data. For example, terms like "makeup" are frequently associated with women, while professions such as "security guard" are often linked with men. These associations reflect demographic biases and social stereotypes commonly embedded in pre-trained models, leading to skewed outputs where certain activities or professions are disproportionately tied to specific genders. This bias can reinforce harmful stereotypes and limit the model's ability to generate precision responses across diverse use cases.

To address these biases, algorithms using causal methods and counterfactual data have been developed to select or generate more representative datasets. By modeling causal relationships between language and demographic factors, these methods help mitigate biased associations, such as gender-linked stereotypes in professions. Techniques like counterfactual data augmentation (CDA) flip demographic attributes in sentences, ensuring more balanced training data and promoting fairness in LLM outputs.

\paragraph{Counterfactual Data Augmentation (CDA)~\cite{zmigrod2019counterfactual}.} The paper enhances corpora through counterfactual data augmentation (CDA) by converting gendered sentences in morphologically rich languages (e.g., Spanish and Hebrew) from masculine to feminine forms, and vice versa. The method involves four steps: first, the sentence is parsed into its morphological and syntactic components. Second, an intervention is made by changing the gender of a key noun (e.g., “ingeniero” to “ingeniera”). Third, a Markov random field model is used to infer how the rest of the sentence’s morphological tags (e.g., verbs, adjectives) should be updated to maintain grammatical agreement. Finally, the model reinflects the sentence by adjusting the surrounding words accordingly (e.g., changing “el” to “la” and “experto” to “experta”).
This process ensures that the gender transformation is done without breaking grammaticality, which is crucial for languages with gender agreement. The augmented sentences can then be used to balance corpora and reduce gender stereotypes in NLP models, improving both the fairness and accuracy of language models in handling gendered terms.

\paragraph{Counterfactually Augmented Dataset (CAD)~\cite{Kaushik2020Learning}.} This paper introduces the idea of a counterfactually augmented dataset, in which each example is paired with a manually constructed example with a different label that involves the minimal possible edit to the original example to make that label correct. This approach helps models distinguish between the causally relevant features and the spurious correlations, allowing them to focus on the essential aspects of the data. For pre-trained models, which often inherit spurious correlations from large-scale, uncurated datasets, this method offers a way to refine the model’s understanding by exposing it to examples that directly challenge these misleading associations.

By training on counterfactually augmented data, pre-trained models can be further fine-tuned to become more robust to the spurious patterns typically present in their pre-training corpora. This method provides a targeted mechanism for eliminating irrelevant features, helping the pre-trained models generalize better across different domains and reducing their reliance on artifacts. The paper’s insights highlight how counterfactual data can complement pre-training, guiding models toward a more causally sound representation of the data and improving their robustness in real-world applications.

\paragraph{Commonsense Counterfactual Generation (CCG)~\cite{miao2023generating}.} This paper introduces the idea of generating commonsense counterfactuals for stable relation extraction, where the goal is to mitigate spurious correlations by generating minimally altered examples that conform to commonsense knowledge. The method leverages WordNet for relation expansion and employs a novel intervention-based strategy to accurately identify causal terms. For pre-trained models, this counterfactual augmentation offers a valuable mechanism for refining learned representations by explicitly challenging spurious correlations inherited from pre-training corpora. By training on such augmented data, pre-trained models can reduce their reliance on irrelevant associations, leading to more robust and reliable performance in downstream tasks. 

\paragraph{Counterfactual Explanation Generation (CEG)
\cite{bhattacharjee2023llms}.} This paper presents a counterfactual generation method using large language models (LLMs) to provide causal explanations for black-box text classifiers. The approach follows a three-step pipeline: first, the LLM identifies latent features that contribute to the classifier’s prediction; second, it links those latent features to specific input words; and finally, it generates a counterfactual by making minimal edits to the identified words to flip the model’s decision. The method focuses on maintaining high semantic similarity between the original and counterfactual inputs, ensuring the explanation remains interpretable and causally informative. For pre-trained models, this counterfactual method provides a valuable mechanism for improving robustness and causal understanding. Pre-trained models often rely on spurious correlations learned during the training phase, which can lead to misleading predictions. By generating counterfactual explanations, this method helps refine the model’s focus on causal features rather than superficial patterns.

\vspace{7pt}
In the pre-training stage of large language models (LLMs), incorporating causality and counterfactuals can significantly improve model learning by encouraging the model to focus on genuine causal relationships in the data. Rather than simply learning from correlations present in large, uncurated datasets, counterfactuals allow the model to understand how minimal changes in the input can lead to different outcomes. This helps the model refine its understanding of the underlying causal mechanisms, reduce reliance on spurious correlations, and improve its ability to generalize effectively across various downstream tasks.

\subsection{Causal Foundation Model}
\label{sec:CFM}
Finally, and most importantly, the construction of the Foundation Model is crucial for Large Language Models (LLMs). The majority of current LLMs use the decoder model of Transformer as their foundation model. This choice excels in handling large-scale text generation tasks, particularly in scenarios requiring strong contextual dependency management. The Transformer decoder leverages the self-attention mechanism to capture long-range dependencies, enabling the model to learn complex patterns and linguistic structures from vast corpora, thereby producing high-quality and coherent text outputs. However, one significant issue is that the attention module in the Transformer primarily captures interactions between words, which makes it susceptible to demographic biases and social stereotypes. This raises a fundamental question: how can we design models that not only manage word interactions but also capture causal relationships between entities, as well as between inputs and outputs? A promising solution lies in the development of Causal Attention mechanisms within the Transformer architecture, a concept that has been explored previously in the domain of computer vision~\cite{wang2021causal,yang2021causal}.

\paragraph{Causal Attention Module (CaaM)~\cite{wang2021causal}.} This paper focuses on eliminating confounding factors in visual recognition tasks through causal intervention. In the Causal Attention Module (CaaM), confounders are identified in a self-supervised way. The fundamental idea is based on the causal intervention formula: $ P(Y \mid do(X)) = \sum_{t \in T} P(Y \mid X, t) P(t) $, where $ P(Y \mid X, t) $ is the prediction in each partitioned environment $ t $, and $ P(t) $ is the environment’s probability. This formula removes bias caused by confounders by adjusting for different environments. An improved version for more complex cases, where both confounders $ S $ and mediators $ M $ are present, is given as $ P(Y \mid do(X)) = \sum_{s \in S} \sum_{m \in M} P(Y \mid X, s, m) P(m \mid X) P(s) $, ensuring that mediators are treated separately from confounders. For training, CaaM uses an adversarial approach where a minimization objective ensures robust causal features are learned: $ \min_{A, A', f, g, h} XE(f, \tilde{x}, D) + IL(g, A(x), T_i) + XE(h, A(x), D) $, and a maximization objective updates the data partitions to better capture confounders: $ \max_{\theta} IL(h, A(x), T_i(\theta)) $. Through these formulas, CaaM implements a causal attention mechanism in visual recognition tasks, improving model robustness when handling biased data. 

\paragraph{Causal Attention (CATT)~\cite{yang2021causal}.} This paper introduces Causal Attention (CATT), a novel attention mechanism designed to remove confounding effects in vision-language models. The approach is based on the front-door adjustment, i.e., $P(Y|do(X)) = \sum_{z} P(Z = z|X) \sum_{x} P(X = x) P(Y|Z = z, X = x)$, which estimates the causal effect of input $X$ on output $Y$ via a mediator $Z$, helping to address spurious correlations in data. In this formulation, $P(Z = z|X)$ is the probability of selecting $Z$ given $X$, which is calculated by In-Sample Attention (IS-ATT), $P(X = x)$ represents Cross-Sample Sampling (CS-ATT) from other examples, and $P(Y|Z = z, X = x)$ models the final prediction using the sampled values. To efficiently implement the Causal Attention mechanism and reduce the computational cost of sampling, the paper introduces the Normalized Weighted Geometric Mean (NWGM) approximation:
\begin{equation}
    P(Y|do(X)) \approx \text{Softmax}[g(\hat{Z}, \hat{X})], \qquad
    \hat{Z} = \sum_{z} P(Z = z|h(X)) z \qquad
    \hat{X} = \sum_{x} P(X = x|f(X)) x 
\end{equation}
where, $g(Z, X)$ represents the predictive function modeled by a neural network, and $\hat{Z}$ and $\hat{X}$ are the estimates for In-Sample and Cross-Sample sampling, respectively. The variables $h(X)$ and $f(X)$ are embedding functions for transforming the input $X$ into query sets for IS-Sampling and CS-Sampling. The NWGM approximation absorbs the outer sampling operations, ensuring that only one forward pass is needed through the network. 

These methods offer valuable insights for pre-trained language models (PLMs), encouraging the disentanglement of causal language patterns from spurious correlations within data. This approach has the potential to improve model robustness, particularly when handling out-of-distribution or biased text inputs. Building on these discussions, recent research has begun to examine the interplay between causality and attention mechanisms in LLMs~\cite{rohekar2024causal}. For instance, Zhang et al.~\cite{zhang2024towards} have introduced the Causal Foundation Model, which represents a significant advancement in this area.

\paragraph{Causal Interpretation of Self-Attention~\cite{rohekar2024causal}.} This paper propose a causal interpretation of the self-attention mechanism used in Pre-Trained Transformer models. They interpret the self-attention mechanism as a way to estimate a Structural Causal Model (SCM) for a given input sequence of symbols (tokens). The relationship between endogenous variables in the SCM is defined as $X_i \leftarrow f_i(\text{Pa}_i, U_i)$, where $ X_i $ is an endogenous variable, $ \text{Pa}_i $ represents its direct causes (parents), and $ U_i $ is an exogenous noise term. In the case of a linear-Gaussian SCM, the relationship is expressed as $X = GX + \Lambda U$, where $ G $ is the weight matrix, and $ \Lambda $ is a diagonal matrix for the exogenous noise. Then, they found that the attention mechanism in Transformers can be understood through the attention matrix $ A $. The covariance $C_Z$ and attention $A$  can be computed as:
\begin{equation}
C_Z = A A^\top, \qquad A = \text{softmax}(Y W_Q W_K^\top Y^\top)
\end{equation}
where $ Y $ represents the input embeddings, and $ W_Q $ and $ W_K $ are learnable weight matrices. The output of the attention mechanism is $Z = AV$ with $ V = Y W_V $ being the value matrix. The core insight is that the covariance matrix of the output embeddings $ Z $ is given by $C_Z = A A^\top$, which can be directly linked to the covariance of input $X$ in SCM:
\begin{equation}
C_X = (I - G)^{-1} \Lambda C_U \Lambda^\top (I - G)^{-1 \top}
\end{equation}
Thus, the attention matrix $A$ is analogous to the causal dependencies encoded in $ (I - G)^{-1} \Lambda $, making self-attention a mechanism that captures causal relationships between input tokens. Finally, the paper introduces the Attention-Based Causal Discovery (ABCD) algorithm, which leverages the attention matrix to learn the causal graph underlying the input sequence. This enables pre-trained Transformers to perform zero-shot causal discovery by interpreting the relationships between input tokens as causal dependencies. Additionally, the paper introduces a method for visualizing these causal relationships between tokens, offering a clearer interpretation of how they influence each other. This approach can further enhance our understanding of language models and serve as a method for improving the development of foundational models.

\paragraph{Causal Inference with Attention (CInA)~\cite{zhang2024towards}.} This paper proposes a Causal Foundation Model aimed at performing causal inference tasks such as treatment effect estimation using a transformer-based architecture. They introduce the method Causal Inference with Attention (CInA), which leverages the self-attention mechanism to estimate optimal covariate balancing weights, a key step in causal inference. This approach allows for zero-shot causal inference, enabling models to estimate treatment effects on unseen data without retraining. 
The main theoretical insight presented is the primal-dual connection between self-attention and optimal covariate balancing. The attention mechanism, represented as $\text{softmax}\left( \frac{Q K^\top}{\sqrt{D}} \right) V$, is shown to be analogous to the solution of a Support Vector Machine (SVM) problem, where weights $\alpha$ from SVM are linked to attention outputs. This duality allows for efficient causal effect estimation by utilizing attention mechanism to compute balancing weights.

The proposed method achieves zero-shot inference by generalizing across multiple datasets through self-supervised learning. Once trained, the model can compute treatment effects on new datasets without needing to re-optimize. The loss function used for training is based on the following regularized hinge loss:

\begin{equation}
    L_\theta(D) = \frac{\lambda}{2} \left\| \sum_{j=1}^N \frac{v_j}{h(X_j)} \phi(X_j) \right\|^2 + \left[ 1 - W \cdot \left( \text{softmax}(KK^\top / \sqrt{D}) V + \beta_0 \right) \right]^+
\end{equation}
This formulation bridges the principles of causal inference with modern deep learning architectures, marking a significant step toward developing foundation models capable of causal reasoning. The paper explores potential enhancements to pre-trained language models (LLMs) by embedding causal inference mechanisms, enabling these models to transition from correlational to causal reasoning. While traditional LLMs excel at pattern recognition in large datasets, they often fall short in tasks requiring a nuanced understanding of causal relationships—particularly crucial in fields such as healthcare and policy-making. By integrating causal inference techniques, such as Causal Inference with Attention (CInA), LLMs could demonstrate improved generalizability across diverse domains, enhance decision-making accuracy, and provide more interpretable outcomes.

\vspace{7pt}
Furthermore, some works explore integrating causal graphs with language models. \cite{zhao2017constructing,phatak2024narrating} explore methods for constructing and narrating causal graphs. \cite{lu2021kelm} converts knowledge graphs into natural language sentences to augment the pre-training corpus of LLMs. \cite{pan2024unifying} aligns textual information and KG subgraphs in a unified pre-training framework, leading to improved contextual understanding and reasoning capabilities. These approaches potentially reveal how we can leverage causal knowledge graphs to further enhance pre-training models. We will defer some future exploration reactions to Section \ref{sec:future}.

\section{Causality for Fine-Tuning}

To make a pre-trained foundation model useful for both specific and general tasks, fine-tuning is essential. In supervised fine-tuning (SFT), the model is refined using labeled data to suit particular tasks. Although modern large language models (LLMs) can often handle tasks without the need for fine-tuning, this process still proves beneficial when optimizing for task-specific or data-specific requirements. Both fine-tuning and pre-training share common elements~\cite{miao2023generating,bhattacharjee2023llms,zhou2023causal}, such as feature extraction, and can incorporate advanced methods like causal feature extraction and counterfactual data augmentation. The main distinctions between the two, however, lie in the scale of the training corpus and the focus on specific tasks. In this section, we will review several causal techniques that have been effectively applied during the fine-tuning stage. These methods aim to enhance the model’s ability to generalize by focusing on the underlying causal relationships in the data, ensuring that fine-tuning goes beyond mere correlation to capture deeper, task-relevant insights~\cite{zhang2024label,zheng2023preserving,chen2023disco,feder2024causal,gat2023faithful}.

\paragraph{Label-aware Debiased Causal Reasoning Network (LDCRN)~\cite{zhang2024label}.} In the Natural Language Inference (NLI) task, this paper proposes the Label-aware Debiased Causal Reasoning Network (LDCRN), which employs causal reasoning to distinguish between genuine and spurious correlations in NLI datasets. By utilizing a causal graph, the method captures interactions between input sentences, labels, and biases, ensuring that the model learns more robust and generalizable features.
LDCRN quantifies the overall causal effect of input data on the inference label. The total effect is given by $TE = Y_{p,h,r,c} - Y_{p^*,h^*,r^*,c^*}$, where $Y_{p,h,r,c}$ represents the model’s prediction using the premise ($p$), hypothesis ($h$), and their interactions ($r$), with $c$ capturing spurious correlations. This total effect reflects both genuine and biased influences. To remove the biases, the label-aware biased module computes the Controlled Direct Effect (CDE) as $CDE = Y_{p^*,h,r^*,c} - Y_{p^*,h^*,r^*,c^*}$, identifying spurious correlations based on the label information. Then the debiased inference is obtained by subtracting the CDE from the TE, resulting in $TIE = Y_{p,h,r,c} - Y_{p^*,h,r^*,c}$. This approach ensures that the model focuses on genuine causal relationships between premises and hypotheses, improving generalization and reducing reliance on biased patterns in the data. LDCRN differs from traditional fine-tuning methods by using causal techniques to remove bias in downstream tasks, rather than simply relying on pure fine-tuning a pre-trained model.

\paragraph{Causal Effect Tuning (CET)~\cite{zheng2023preserving}.} This paper revolves around understanding and addressing catastrophic forgetting during fine-tuning of pre-trained language models for commonsense reasoning tasks. The authors propose the Causal Effect Tuning (CET) method, which applies causal inference to identify and retain valuable pre-trained knowledge during the fine-tuning process. CET frames the fine-tuning process using causal graphs to uncover the missing causal effects from pre-trained data that lead to forgetting.

A key formula introduced in the paper describes the causal effect of pre-trained data $P$ on model predictions $\hat{Y}$ during fine-tuning: $\text{Effect}_P = P(\hat{Y} = \hat{y} | do(P = p)) - P(\hat{Y} = \hat{y} | do(P = 0))$.This formula quantifies the difference in predictions between models fine-tuned with and without pre-trained data, highlighting the potential loss of knowledge during standard fine-tuning. CET mitigates this by using K-nearest neighbors (KNN) in the feature space to capture colliding effects: $\text{Effect}_P^{(i)} = \sum_{k=0}^{K} P(\hat{Y}^{(i)} | X = x^{(i,k)}) W_P(x^{(i)}, x^{(i,k)})$. This equation preserves pre-trained knowledge by weighing the influence of similar samples in the hidden space. The overall unified objective combines vanilla fine-tuning and causal objectives to balance knowledge retention and learning from new data: $\max \sum_{i \in S_T} \sum_{k=0}^{k_i} P(\hat{Y}^{(i)} | X = x^{(i,k)}) W_P(x^{(i)}, x^{(i,k)}) + \sum_{i \in S_{NT}} P(\hat{Y}^{(i)} | X = x^{(i)})$. CET outperforms existing methods, effectively preserving commonsense knowledge and improving model performance on downstream tasks.

\paragraph{Distilling Counterfactuals (DISCO)~\cite{chen2023disco}.}
The paper introduces a method to generate high-quality counterfactual data at scale, aimed at improving robustness and generalization of models in natural language inference (NLI) tasks. The key challenge addressed in the paper is the scarcity of counterfactual data, which is crucial for training models to understand causal relationships. To address this, the authors propose the DISCO (DIStilled COunterfactual Data) approach, which leverages large language models (LLMs) like GPT-3 to automatically generate counterfactuals and then uses a task-specific teacher model to filter and distill the generated data, ensuring high-quality examples for training. The core of the DISCO method involves two key steps. First, the LLM is used to generate diverse perturbations by modifying specific spans of text within the input data. The perturbations aim to flip the original labels (e.g., from "Entailment" to "Contradiction") by altering critical parts of the input. The second step involves filtering these generated perturbations using a specialized teacher model that ensures the new label is appropriate and causally consistent with the perturbed input. This process is formalized through the calculation of a distributional shift: $\Delta l' = p(l'|P', H) - p(l'|P, H)$, where, $P'$ represents the perturbed premise, $H$ is the hypothesis, and $l'$ is the new label. The shift in prediction probability $p$ is used to determine whether the perturbation successfully flips the label as intended.

DISCO is applied to the NLI task, where it shows significant improvements in robustness and generalization compared to models trained without counterfactual data. The generated counterfactual data is shown to be diverse and effective, yielding improvements in both counterfactual accuracy and sensitivity. These metrics are defined as: Counterfactual accuracy (CAcc) measures the consistency of the model in correctly predicting both the original and counterfactual examples:$CAcc = \frac{1}{K} \sum_{k=1}^{K} 1 \left((\hat{l}_k = l_k) \land (\hat{l}'_k = l'_k)\right)$; Counterfactual sensitivity ($\delta_s$) quantifies how much the model's predictions shift between the original and perturbed examples:$\delta_s = \frac{(p(\hat{l}'|x') - p(\hat{l}'|x)) + (p(\hat{l}|x) - p(\hat{l}|x'))}{2}$. Overall, the paper demonstrates that DISCO-generated counterfactuals lead to improved performance on several NLI benchmarks, showing the potential of large language models for generating high-quality, diverse training data without requiring extensive human annotation.

\paragraph{Causal-Structure Driven Augmentations \cite{feder2024causal}.} The paper proposes a method to enhance the generalization of text classifiers, especially for out-of-distribution (OOD) scenarios, by addressing issues of spurious correlations through causal structure-based counterfactual data augmentation. The authors argue that language models often rely on spurious correlations, which limits their ability to generalize across different distributions or environments. To solve this, the paper introduces a framework that incorporates causal knowledge into data augmentation, enabling the generation of counterfactual text instances that simulate interventions on confounding variables, such as spurious features present in the training data.

The core idea of the method is to simulate alternative scenarios using counterfactual data generation, driven by knowledge of the underlying causal structure of the dataset. For example, in clinical diagnosis from medical narratives, the authors simulate what would happen if a different caregiver wrote the same clinical notes. By training models on these counterfactual instances, the classifier becomes less dependent on spurious correlations like the writing style of a specific caregiver, leading to improved OOD generalization. The paper shows through experiments that this causal augmentation approach results in superior performance compared to traditional re-weighting methods and invariant learning techniques.

Formally, the method can be understood as aiming to minimize the risk $\mathbb{E}_{P}(L(f(X), Y))$ under a distribution $P$ where the spurious feature is deconfounded from the label. To achieve this, the augmented data is generated based on the causal structure, producing examples as if the confounding factor had been intervened upon, thus breaking the spurious correlations. The approach is backed by both theoretical guarantees on sample complexity and empirical results showing improved performance across multiple domains, such as clinical text classification and restaurant review tasks.

\paragraph{LLM4Causal~\cite{jiang2023large}} The LLM4Causal model is an end-to-end large language model designed for causal decision-making tasks by combining fine-tuning and tool integration techniques. It is fine-tuned on two custom datasets, Causal-Retrieval-Bench and Causal-Interpret-Bench, to classify causal queries, extract relevant parameters, and interpret causal outputs. The model identifies causal tasks like causal graph learning (CGL), average treatment effect estimation (ATE), and heterogeneous treatment effect estimation (HTE) from user queries, extracting key variables such as treatments, outcomes, and mediators. By fine-tuning on these datasets, LLM4Causal can handle structured data, process user inputs in a structured JSON format, and produce well-defined results that address specific causal questions. Once the causal task is identified, LLM4Causal integrates with existing causal tools like CausalML or CausalDM to perform causal inference, automatically selecting the right tool and executing the analysis based on the task. The model then interprets the numerical or graph-based results into human-readable summaries, making the insights accessible to general users. Its modular approach allows flexibility in incorporating new causal tools and algorithms. Extensive evaluations, including ablation studies, show that LLM4Causal outperforms baseline models in terms of query handling, task relevance, and output accuracy, making it a powerful tool for causal reasoning tasks.

\paragraph{Matching \cite{gat2023faithful}.} This paper introduces two novel approaches for explaining black-box NLP models through causal counterfactuals. The first method involves using large language models (LLMs) to generate counterfactual explanations, prompting them to modify specific text attributes while keeping other confounding variables constant. This allows for a clear understanding of how changes in one concept influence a model’s predictions, thus providing a faithful explanation. While this method shows strong performance, it is computationally expensive, making it less feasible for real-time applications.

To address this limitation, the second approach, known as the matching technique, proposes training a dedicated embedding space guided by an LLM at the training stage. This embedding space is designed to be faithful to a given causal graph and enables efficient matching of text instances that serve as counterfactual approximations. By doing so, the model can estimate causal effects and provide explanations with far less computational overhead, making it a practical alternative to full-scale LLM-based generation.

The paper emphasizes that approximating counterfactuals is crucial for constructing faithful model explanations, as non-causal methods can lead to misleading interpretations. The authors provide theoretical backing for their approach and demonstrate through experiments that both methods—counterfactual generation and matching—surpass traditional explanation techniques in providing model-agnostic, causally faithful insights into NLP systems. While pre-trained large language models are involved in the counterfactual generation process, the core work centers on how these pre-trained models are fine-tuned to generate counterfactuals and improve the matching method.

\vspace{7pt}
These methods use LLMs to improve model explainability and generalization~\cite{miao2023generating,bhattacharjee2023llms,zhou2023causal,zhang2024label,zheng2023preserving,chen2023disco,feder2024causal,gat2023faithful}. In the counterfactual generation method, LLMs are prompted to modify text, and these modifications can be further enhanced by fine-tuning the model, such as by generating additional useful training data. In the matching technique, the fine-tuned embedding space is used to find the most similar counterfactual instances to effectively explain the model’s predictions. Fine-tuning not only improves the interpretability of LLMs but also helps it capture high-level conceptual relationships, making it more accurate and reliable for complex tasks.

\section{Causality for Alignment}

AI Alignment is the process of guiding AI systems to behave in ways that align with human goals, preferences, and ethical standards. This is especially important because large language models (LLMs), while pre-trained for tasks like predicting the next word in a sentence, can unintentionally produce harmful, toxic, misleading, or biased content. By aligning AI systems with human values, we can reduce these risks and ensure that the models generate safer, more reliable, and ethically sound outputs. Several techniques have been developed to achieve alignment, including Proximal Policy Optimization (PPO)~\cite{schulman2017proximal}, a reinforcement learning method designed to improve the stability and efficiency of policy updates, often used to optimize models in the alignment process. Reinforcement Learning from Human Feedback (RLHF)~\cite{christiano2017deep} adjusts models based on human evaluations of their outputs, guiding them to generate responses more aligned with human preferences. More recently, Direct Preference Optimization (DPO)~\cite{rafailov2023direct} has been introduced to directly tailor models to better match human preferences without the complexities of reinforcement learning.

\paragraph{MoCa~\cite{nie2024moca}.}
This paper evaluates the alignment of large language models (LLMs) with human judgments in scenarios requiring causal and moral reasoning. The authors collected a dataset from cognitive science literature and designed a challenge set where small changes in text lead to significant shifts in human judgments. They examined various LLMs, including models fine-tuned with reinforcement learning from human feedback (RLHF), to assess whether the models align with human intuitions. Recently, while larger models show improved alignment, they often weigh causal and moral factors differently than humans. Using Average Marginal Component Effect (AMCE) analysis, $\Delta(j) = \sum_{j=1}^{J} \sum_{n=1}^{N} \sum_{k=1}^{K} \frac{1{T_{kj} = 1} Y_{jnk}}{1{T_{kj} = 1}} - \sum_{j=1}^{J} \sum_{n=1}^{N} \sum_{k=1}^{K} \frac{1{T_{kj} = 0} Y_{jnk}}{1{T_{kj} = 0}}$, the authors uncovered key differences between model and human judgments. For example, LLMs tend to give greater weight to abnormal events when determining causality, and exhibit varying sensitivities to moral factors such as personal force or the inevitability of harm. These findings demonstrate the value of carefully structured challenge datasets for evaluating model performance beyond aggregate metrics, highlighting the need for further refinement in training methods to ensure better alignment between LLMs and human reasoning.

\paragraph{Counterfactual DPO~\cite{butcher2024aligning}.} This paper introduces a method for aligning large language models (LLMs) using Direct Preference Optimization (DPO) with counterfactual prompting, as an alternative to the traditional Reinforcement Learning from Human Feedback (RLHF). RLHF is resource-intensive and relies on human evaluators to rank model outputs. In contrast, DPO directly optimizes the model by using counterfactual prompts to guide the LLM's behavior, where specific desired styles are preferred and undesired ones are minimized. This eliminates the need for human intervention while maintaining alignment with ethical and responsible behaviors.

The DPO approach is represented by a loss function $L_{DPO}(\pi_{\theta}; \pi_{sft}) = -E(x, y_w, y_l) \sim D \left[ \log \sigma \left( M(x, y_w, y_l) \right) \right]$, where $M(x, y_w, y_l)$ defines the log-margin between preferred ($y_w$) and rejected ($y_l$) responses under prompt $x$. By maximizing the likelihood of preferred responses and reducing undesirable ones, the model is fine-tuned through maximum likelihood estimation without requiring human evaluators. In summary, the paper demonstrates that counterfactual DPO effectively reduces hallucinations, mitigates biases, and enables LLMs to ignore inappropriate instructions. The method provides a more scalable and resource-efficient way to align LLMs with human expectations, promoting ethical AI behavior across diverse applications.

\paragraph{Causal Preference Optimization~\cite{lin2024optimizing}.} This paper introduces two innovative methods for optimizing large language models (LLMs) based on human preferences: Causal Preference Optimization (CPO) and its enhanced version, Doubly Robust Causal Preference Optimization (DR-CPO). The key idea behind these methods is to treat LLM alignment as a causal inference problem, ensuring that the model learns to generate texts that cause desired outcomes in users. Unlike traditional methods such as Reinforcement Learning from Human Feedback (RLHF), which rely on labor-intensive paired datasets, CPO operates on direct outcome datasets, where each sample consists of a text and a corresponding numerical measure of user response.

The CPO method seeks to maximize the likelihood of generating texts that lead to preferred outcomes by using importance weighting. The core value function for CPO is expressed as $ V(f) = E_{X \sim P_f} [E_{Y(·) \sim G} [Y(X)]] $, where $P_f$ is the distribution of texts generated by the model and $G$ is the potential outcomes framework. CPO directly adjusts the model to increase the probability of generating texts that align with human preferences, ensuring the model learns causal, not merely correlational, relationships between text and outcome. DR-CPO builds on CPO by introducing a doubly robust framework, which combines importance weighting with outcome modeling to further improve the model’s robustness. The value function for DR-CPO is given by: 
\begin{equation}
    V(f) = E_{X \sim P_R} \left[ E_{Y \sim P_R^y} \left[ \frac{P_f(X)}{P_R(X)} (Y - g(X)) \right] \right] + E_{X \sim P_f} [g(X)].
\end{equation}
This formulation ensures that even if either the outcome model $g(X)$ or the importance weights are misspecified, the method remains unbiased. By combining randomized and non-randomized data sources, DR-CPO can reduce variance while maintaining strong bias guarantees, making it more effective in handling real-world, confounded datasets.
Both CPO and DR-CPO represent significant advancements in optimizing LLMs for human preferences. They provide scalable, efficient alternatives to RLHF by leveraging causal inference techniques, allowing models to align more closely with desired human behaviors across various applications while mitigating common issues like bias and confounding.

\vspace{7pt}
Recent advances in AI alignment have increasingly focused on incorporating causality to ensure that large language models (LLMs) not only mirror human preferences but also understand the underlying causal relationships that drive those preferences. Methods like Causal Preference Optimization (CPO) and Doubly Robust Causal Preference Optimization (DR-CPO) treat alignment as a causal inference problem, enabling models to learn how their outputs causally influence user responses rather than relying on mere correlations. Similarly, approaches such as counterfactual Direct Preference Optimization (DPO) use counterfactual prompts to guide LLMs toward desired behaviors without extensive human intervention. Evaluations with carefully designed datasets, like those in the MoCa study, reveal that while LLMs are improving, they often weigh causal and moral factors differently than humans do. By integrating causal reasoning into alignment techniques, future research aims to develop AI systems that better reflect human judgment, handle complex moral considerations, and ultimately produce outputs that are safer, more reliable, and ethically aligned with human values.

\section{Causality for Inference}
\begin{figure}[t]
\begin{center}
\includegraphics[width=1.0\linewidth]{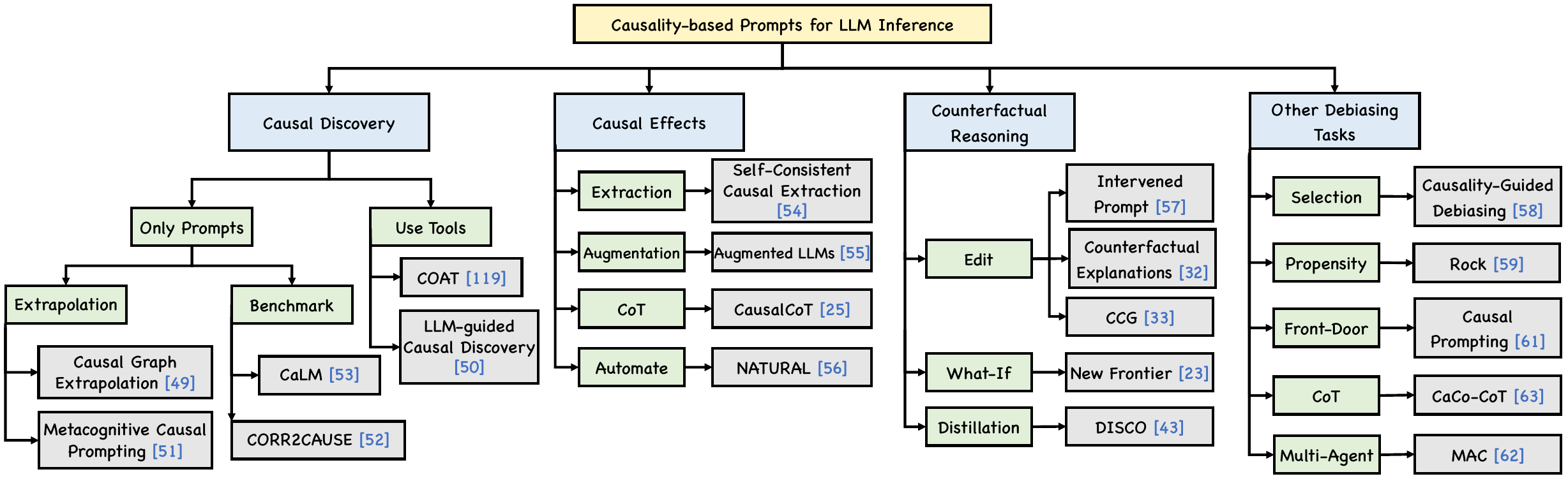}
\end{center}
\vspace{-12pt}
\caption{Categorization of Causality-based Promts for LLM Inference.}
\label{fig:inference}
\end{figure}

Natural language serves as a repository of knowledge and information, primarily functioning as a tool for communication rather than a medium for thought~\cite{fedorenko2024language}. After being trained on large-scale human language networks, LLMs can recite knowledge to respond to various language tasks, yet they still do not know how to apply this knowledge or think independently. Therefore, human intervention is required to provide ‘thoughtful’ prompts that guide the LLMs, shaping their responses to ensure the integration of relevant knowledge and reasoning. These processes are called 'prompt engineering'~\cite{giray2023prompt,white2023prompt}. To enhance the reliability and depth of LLM responses, recent research has proposed designing causal prompts or causal chain-of-thoughts, which activate LLMs to recall causal knowledge and integrate it into their responses for more accurate and insightful answers\cite{bhattacharjee2023llms,antonucci2023zero,vashishtha2023causal,ohtani2024does,abdali2023extracting,tan2023causal,romanou2023crab,zhang2023causal}. In Figure \ref{fig:inference}, we categorize these studies into four distinct groups, each focusing on causality-based prompts for LLM inference in various tasks. These include carefully crafted prompts used for Causal Discovery, Causal Effect Estimation, Counterfactual Reasoning, and debiasing efforts. Detailed examples of these prompts are provided in Tables \ref{tab:prompt1}-\ref{tab:prompt4}.

\subsection{Causal Discovery}

%%%%%%%%%%%%%%%%%%%%%%%%%%%%%%%%%%%
%%%%%%%%%%%%%%%%%%%%%%%%%%%%%%%%%%%
%%%%%%%%%%%%%%%%%%%%%%%%%%%%%%%%%%%
%%%%%%%%%%%%%%%%%%%%%%%%%%%%%%%%%%%

As discussed earlier, LLMs can recall causal knowledge encountered during their training on vast internet-scale data, including causality knowledge in causa literature~\cite{orhan2023recognition}. Through exposure to technical books and academic papers, LLMs can store and retrieve causal relationships or causal graphs from these sources~\cite{kiciman2023causal, ashwani2024cause, ma2024causal}. While LLMs can generate responses that reference such causal discoveries, this process occurs largely without genuine comprehension or reasoning about the recalled information~\cite{ma2024causal}. This motivates researchers to develop more specific prompts that can trigger LLMs to retrieve relevant causal information, thereby facilitating the integration of causal reasoning into response generation~\cite{chen2024causal,jin2024can,liu2024discovery,antonucci2023zero,vashishtha2023causal,ohtani2024does}. Various approaches, such as those proposed in CaLM~\cite{chen2024causal} and CORR2CAUSE~\cite{jin2024can}, provide new benchmarks and explore how prompting and model fine-tuning can enhance LLMs’ capacity for causal discovery and inference. Other frameworks, including Causal Graph Extrapolation and Metacognitive Causal Prompting, demonstrate the extraction of causal relations from natural language texts using large language models (LLMs) without requiring explicit training data. Additionally, methods like COAT~\cite{liu2024discovery} and LLM-guided Causal Discovery~\cite{vashishtha2023causal} integrate LLM outputs with traditional causal discovery techniques to improve accuracy and reasoning. Below, we present a detailed overview of these methods.

\paragraph{CaLM~\cite{chen2024causal}.}  To enhance the causal reasoning abilities of LLMs, Chen et al. introduce CaLM, a causality benchmark for assessing the capabilities of 28 leading language models on a core set of 92 causal targets, 9 adaptations, 7 metrics, and 12 error types~\cite{chen2024causal}. This paper establishes a structured framework with four core components—causal target, adaptation, metric, and error—to systematically assess how well LLMs understand and apply causal relationships. By creating a dataset of over 126,000 samples in English and Chinese across natural, symbolic, and mathematical texts, they conduct an in-depth analysis of factors affecting model performance and identify interdependencies between different evaluation dimensions. Taking Causal Discovery as an example, they propose various prompting methods, i.e., (1) Basic Prompt, (2) Adversarial Prompt, (3) Chain-of-Thought (CoT), (4) In-context Learning, and (5) Explicit Function—to address four question types in the 92 causal targets: Binary classification, Choice selection, Open-ended generation, and Probability calculation.

\paragraph{CORR2CAUSE~\cite{jin2024can}.} This paper introduces a new task, CORR2CAUSE, to test whether large language models (LLMs) can infer causal relationships from correlational statements. The authors propose six hypothesis templates for pairwise causal discovery: Is-Parent, Is-Child, Is-Ancestor, Is-Descendant, Has-Confounder, and Has-Collider. These templates help the model determine causal relations between variables based on statistical correlations. The dataset constructed includes over 200K samples, and the experiments show that while LLMs can perform better with fine-tuning, their out-of-distribution generalization remains limited.

\paragraph{Zero-shot Causal Graph Extrapolation~\cite{antonucci2023zero}.} The paper presents a method for extracting causal relations from natural language texts using large language models (LLMs) without the need for explicit training data. The authors employ an iterative pairwise query approach, leveraging LLMs to infer causal graphs directly from textual data, particularly in scientific and medical domains. The results from evaluating biomedical abstracts show that LLMs can accurately identify causal relations with a high recall, although they sometimes struggle to distinguish direct from indirect causal connections, leading to false positives. This approach highlights the potential of LLMs in automating causal discovery in large volumes of scientific literature.

\paragraph{Metacognitive Causal Prompting~\cite{ohtani2024does}.} The paper investigates whether metacognitive prompting enhances the causal reasoning capabilities of large language models (LLMs). Through a series of experiments, the authors compare the performance of LLMs on causal inference tasks, such as identifying sufficient and necessary causes, both with and without metacognitive prompts. The results show that while metacognitive prompting improves performance in certain scenarios, it is not always effective, and in some cases, LLMs may appear to conduct deeper reasoning without genuinely doing so. While metacognitive prompting holds potential, further refinement is needed to ensure that LLMs perform genuine metacognitive inference.

\paragraph{COAT~\cite{liu2024discovery}.} This paper develops a novel framework called COAT (Causal representatiOn AssistanT), which leverages Large Language Models (LLMs) for discovering high-level causal factors from unstructured data. Unlike traditional causal discovery approaches that rely heavily on manually annotated, structured data, COAT uses LLMs to propose potential causal factors from raw observational data. These factors are then processed and structured using a causal discovery algorithm, such as FCI (Fast Causal Inference), to uncover causal relationships. The COAT framework iteratively improves the causal factor proposals by generating feedback based on the results of causal analysis, ensuring more accurate and reliable causal discovery. 

\paragraph{LLM-guided Causal Discovery~\cite{vashishtha2023causal}.}This paper introduces a method to improve causal inference by leveraging large language models (LLMs) to determine causal order among variables, which can be easier to elicit than the full causal graph. The authors propose a triplet-based prompting technique to enhance LLMs’ ability to infer causal relationships, outperforming previous pairwise approaches. Additionally, they combine LLM outputs with existing causal discovery algorithms, such as constraint-based and score-based methods, to improve accuracy. Experiments show that LLM-augmented approaches provide more accurate causal orders, demonstrating the potential of LLMs to significantly enhance causal discovery processes.

\vspace{7pt}
Recent progress in utilizing LLMs for causal discovery has yielded promising results through the application of advanced techniques, such as refined prompting strategies and the integration of LLM outputs with traditional causal discovery methods. Studies like CaLM\cite{chen2024causal}, CORR2CAUSE\cite{jin2024can}, and COAT\cite{liu2024discovery} illustrate how carefully designed prompts can enable LLMs to infer causal relationships from large and diverse datasets. Additionally, frameworks such as LLM-guided Causal Discovery\cite{vashishtha2023causal} and Zero-shot Causal Graph Extrapolation~\cite{antonucci2023zero} contribute to improving the precision and reliability of causal inference.

\subsection{Causal Effects}
Another line of research focuses on utilizing prompting techniques for causal effect estimation, aiming to improve LLMs’ ability to infer causal effects when an intervention or policy is applied to specific outcomes across various domains~\cite{abdali2023extracting, pawlowski2023answering, jin2024cladder, dhawan2024end}. These efforts seek to guide LLMs in accurately estimating the impact of interventions by leveraging structured prompts and reasoning processes. First, to process the unstructured text data, Self-Consistent Causal Extraction~\cite{abdali2023extracting} introduces a method that uses in-context learning and a self-consistency mechanism to extract key causal variables and construct causal graphs from unstructured data. Augmented LLMs~\cite{pawlowski2023answering} enhance this process further by using context and tool augmentation to refine causal outputs. Meanwhile, CausalCoT~\cite{jin2024cladder} employs a chain-of-thought prompting strategy, guiding LLMs to generate accurate causal inferences by breaking down complex queries. Finally, the NATURAL framework~\cite{dhawan2024end} automates causal effect estimation from unstructured text data by applying traditional causal inference techniques, providing a scalable and accurate solution. Below, we provide a detailed description of these methods.

\paragraph{Self-Consistent Causal Extraction~\cite{abdali2023extracting}.}
The paper proposes a method for extracting causal variables (treatment, outcome, [confounders]) and constructing causal graphs using large language models (LLMs) with in-context learning. The goal is to leverage these causal graphs for topic modeling. To reduce hallucinations, the authors introduce a modified self-consistency approach that generates multiple reasoning paths and selects the most consistent one. This method identifies key causal variables from user feedback, allowing for the creation of causal graphs and generating actionability scores based on the extracted insights. By combining chain-of-thought prompting and feedback classification, the framework helps automate the process of diagnosing user-reported issues and improves causal analysis without prior domain knowledge.

\paragraph{Augmented LLMs~\cite{pawlowski2023answering}.} The approach presented in \cite{pawlowski2023answering} utilizes basic text completion and introduces two distinct augmentation strategies to enhance LLMs’ ability to respond to causal queries. The first method, context augmentation, involves supplementing the LLM with additional operational capabilities outside of a causal expert system. The second method, tool augmentation, allows the LLM to use external Python tools to refine and format the output of the causal expert system, making it easier for the LLM to interpret.

\paragraph{CausalCoT~\cite{jin2024cladder}.} This paper introduces a chain-of-thought prompting strategy inspired by the Causal Inference (CI) engine, which guides LLMs to first identify key components from the question, including the causal graph, causal query, and relevant “data” (e.g., conditional or interventional do-probabilities). With these elements extracted, the LLM is prompted to generate accurate causal inferences, simulating the reasoning process necessary to solve causal questions effectively.

\paragraph{NATURAL~\cite{dhawan2024end}.} This paper introduces a NATURAL framework, which uses large language models (LLMs) to automate the estimation of causal effects from unstructured text data, such as social media posts or clinical reports. This approach represents a significant advancement in causal inference, as it leverages the vast amounts of freely available text data to estimate causal effects without the need for labor-intensive data curation or randomized controlled trials (RCTs). The NATURAL estimators extract structured variables (such as treatment, outcome, and covariates) from unstructured text using LLMs, and then apply classical causal inference techniques like inverse propensity score weighting and outcome imputation to estimate the average treatment effect (ATE). A key contribution of this framework is its ability to process unstructured, observational data to yield causal effect estimates that are within 3 percentage points of ground truth ATEs from clinical trials, providing a cost-effective and scalable alternative to traditional methods. This innovative method greatly expands the applicability of causal inference to new domains, enabling researchers to utilize real-world evidence from unstructured sources while ensuring consistency with causal assumptions.

\vspace{7pt}
Leveraging prompting techniques for causal effect estimation offers significant potential to enhance LLMs’ ability to inform decision-making in high-risk scenarios, such as medical treatment planning and stock market investments. Methods like Self-Consistent Causal Extraction~\cite{abdali2023extracting}, Augmented LLMs~\cite{pawlowski2023answering}, CausalCoT~\cite{jin2024cladder}, and the NATURAL~\cite{dhawan2024end} framework each contribute unique strategies to improve causal inference, from refining the extraction of causal variables to automating analysis from unstructured data. These advancements enable LLMs to generate more accurate and reliable causal estimates, ultimately supporting better decision-making in critical domains where precise outcomes are essential.

\subsection{Counterfactual Reasoning}
There has been a growing interest in exploring the counterfactual reasoning capabilities of large language models~\cite{kiciman2023causal,bhattacharjee2023llms,miao2023generating,chen2023disco,tan2023causal,yu2023ifqa,huang2023clomo,frohberg2022crass}. Numerous works have focused on designing specific datasets to assess these capabilities, such as IfQA~\cite{yu2023ifqa}, Clomo~\cite{huang2023clomo}, and Crass~\cite{frohberg2022crass}, which we will discuss in detail in Section \ref{Sec:Benchmark}. Similarly, several prompt engineering methods have been developed to enhance the counterfactual reasoning abilities of LLMs, providing more robust frameworks for hypothetical scenario analysis. Some approaches explore the effects of intervening on specific nodes within the chain-of-thought (CoT) reasoning process, as seen in Intervened Prompt\cite{tan2023causal}, which evaluates how modifications to reasoning steps influence final outcomes. Other methods, such as Counterfactual Explanations\cite{bhattacharjee2023llms} and Commonsense Counterfactual Generation (CCG)\cite{miao2023generating}, focus on generating meaningful counterfactuals to explain model decisions or adhere to commonsense constraints. Additionally, techniques like DISCO\cite{chen2023disco} generate high-quality counterfactual data at scale to improve model robustness, while works such as New Frontier~\cite{kiciman2023causal} demonstrate the emergent counterfactual reasoning capabilities of LLMs in tasks like causal DAG generation. These diverse approaches highlight the potential of LLMs in advancing counterfactual reasoning, offering new insights into their applications for causal analysis and decision-making.

\paragraph{Intervened Prompt~\cite{tan2023causal}.} The paper explores the counterfactual reasoning of LLMs by intervening on specific nodes in the chain-of-thought (CoT) reasoning process of arithmetic word problems. The authors apply causal abstraction to test whether the intermediate steps generated by the LLMs genuinely influence their final answers. By modifying specific values or reasoning steps and observing the changes in the outcome, they assess whether the CoT structure governs the model’s behavior. This method reveals the extent to which LLMs use CoTs to reason through problems, offering insights into their counterfactual reasoning capabilities.

\paragraph{Counterfactual Explanations~\cite{bhattacharjee2023llms}.} This paper proposes a novel pipeline that leverages large language models (LLMs) to generate counterfactual explanations for black-box text classifiers, contributing significantly to causal reasoning in NLP tasks. The process is structured in three steps. First, the LLM identifies latent, unobserved features in the input text that are associated with the prediction made by the classifier. These latent features might include abstract concepts such as “tone” or “ambiguity.” Second, the LLM is then used to identify the specific words or input features in the text that correspond to the identified latent features. In the final step, the model minimally modifies these input words to generate a counterfactual version of the text—essentially, an edited version that would cause the black-box classifier to flip its prediction. This approach differs from traditional counterfactual generation methods by focusing on latent features rather than surface-level word changes, which allows for more meaningful and causal explanations. The contribution to counterfactual reasoning lies in the pipeline’s ability to preserve the semantic similarity of the input text while generating explanations that reveal what aspects of the text caused the classifier’s decision and how slight modifications would lead to a different outcome. This novel method could be extended to other forms of causal explainability, suggesting broader applications in fields like causal inference and discovery in NLP.

\paragraph{Commonsense Counterfactual Generation (CCG)~\cite{miao2023generating}.} Furthermore, the paper proposes a framework for generating commonsense counterfactuals, called Commonsense Counterfactual Generation (CCG), which introduces an intervention-based strategy and utilizes the knowledge base WordNet to ensure that the counterfactual generation process adheres to both the minimal perturbation requirement and commonsense constraints. In generating counterfactuals, the CCG framework first accurately identifies causal terms and then uses WordNet’s semantic hierarchy to expand the relations between entities. The counterfactuals are generated by minimally modifying the input to flip the label, thus providing clearer decision boundaries for the model. This approach strongly supports counterfactual reasoning.

\paragraph{New Frontier~\cite{kiciman2023causal}.} This paper explores the causal reasoning capabilities of Large Language Models (LLMs) through extensive experiments, focusing on tasks like causal DAG generation, counterfactual reasoning, and token causality. It demonstrates that LLMs outperform traditional methods, with GPT-4 achieving 97\% accuracy in pairwise causal discovery and 92\% in counterfactual reasoning tasks. A key contribution is the detailed examination of how LLMs, particularly GPT-4, excel in inferring causal direction and counterfactual reasoning through emergent capabilities. The paper emphasizes the potential of LLMs to support causal analysis, especially in counterfactual reasoning, and suggests future research on combining LLMs with traditional causal inference techniques for improved reliability in critical applications.

\paragraph{DISCO~\cite{chen2023disco}.} This paper introduces an innovative approach called DISCO (DIStilled COunterfactual Data) to generate high-quality counterfactual data at scale using large language models (LLMs). Their method tackles the challenge of dataset biases by employing a teacher-student model framework to distill reliable counterfactual examples, which improves model robustness in natural language inference (NLI) tasks. First, the method uses a large general LLM, such as GPT-3, to overgenerate perturbations of input sentences. Then, a task-specific teacher model is applied to filter these perturbations, retaining only those that maintain causal consistency with the original task. The key contribution of DISCO lies in its ability to create counterfactual data automatically without relying on expensive human annotation while producing diverse and causally relevant examples. Models trained on this distilled data show significant improvements in robustness and generalization, outperforming baselines on various stress tests and out-of-domain datasets. Notably, DISCO-augmented models achieve higher counterfactual sensitivity and accuracy, demonstrating the method’s ability to enhance models’ reasoning about changes in context that affect predictions.

\vspace{7pt}
Overall, the exploration of counterfactual reasoning in large language models is advancing rapidly, with various methods and datasets designed to assess and enhance their capabilities. From intervening in chain-of-thought reasoning to generating meaningful counterfactual explanations, these techniques demonstrate the growing potential of LLMs in hypothetical scenario analysis and causal reasoning~\cite{kiciman2023causal,bhattacharjee2023llms,miao2023generating,chen2023disco,tan2023causal}. The development of robust prompt engineering methods and the generation of high-quality counterfactual data are key to improving the accuracy and reliability of LLMs in high-impact domains like healthcare and finance. As research in this area continues, LLMs are poised to play an increasingly significant role in supporting decision-making through a deeper understanding of causal relationships and counterfactual scenarios.

\subsection{Other Debiasing Tasks}
Finally, we turn the attention to causal-inspired debiasing prompts in general tasks. Several approaches have been developed to mitigate biases in large language models (LLMs) by leveraging causal reasoning principles. For instance, the Causality-Guided Debiasing Framework~\cite{li2024steering} utilizes causal models to address social biases in LLMs, offering targeted interventions through prompts that discourage biased reasoning while promoting bias-free outputs. Similarly, ROCK~\cite{zhang2022rock} applies causal inference to commonsense reasoning tasks by employing temporal propensity scores to tackle confounding factors, thereby enhancing the robustness of causal reasoning in event prediction. In contrast, Causal Prompting~\cite{zhang2024causal} employs the front-door adjustment from causal inference, refining prompts to decompose the causal relationship between input and output, and reducing bias through contrastive learning. CaCo-CoT~\cite{tang2023towards} focuses on ensuring causal consistency in knowledge reasoning tasks by using a multi-agent system that evaluates answers through both causal and counterfactual lenses to maintain logical soundness. Lastly, MAC~\cite{duong2024multi} integrates multi-agent capabilities for causal discovery, relying on agent debates and statistical methods to identify causal relationships and generate causal graphs. Collectively, these methods illustrate the significant potential of causal-inspired debiasing frameworks to mitigate biases in various general tasks, improving both fairness and accuracy in LLM outputs.

\paragraph{Causality-Guided Debiasing Framework~\cite{li2024steering}.} This paper introduces a causality-guided debiasing framework to address the issue of social bias in large language models (LLMs). The framework is grounded in causal reasoning, focusing on how biases in LLMs’ responses are tied to the data-generating processes during pre-training and the selection mechanisms within the LLM inference process. Their key contribution lies in constructing detailed causal models that outline how biases from the training corpus are incorporated into the LLM’s decision-making process. They propose a series of prompting strategies aimed at both discouraging biased reasoning (e.g., by inhibiting the use of demographic stereotypes) and encouraging bias-free reasoning (e.g., by focusing on relevant non-demographic information). The causal framework also unifies existing debiasing approaches, such as in-context contrastive examples and inhibitive instructions, and offers new ways to guide LLMs towards making unbiased predictions. Empirically, they show that their framework effectively reduces bias across a variety of tasks, including gender and demographic biases, even when applied to closed-source LLMs where direct access to the model’s internal parameters is not possible. This framework makes a significant contribution to counterfactual reasoning by not only identifying how biased associations form but also proposing structured interventions (through prompts) to counteract and mitigate these biases.

\paragraph{Rock~\cite{zhang2022rock}.} This paper develops a novel ROCK framework to apply causal inference principles to commonsense reasoning tasks by leveraging pre-trained language models. ROCK focuses on addressing confounding effects using temporal propensity scores, which are analogous to the propensity scores used in observational studies. The key idea is to estimate the change in the likelihood of one event ($E1$) leading to another event ($E2$) by comparing the probabilities of $P(E1 \prec E2)$ and $P(\neg E1 \prec E2)$, where the difference $\Delta = P(E1 \prec E2) - P(\neg E1 \prec E2)$ captures the causal effect. To mitigate the impact of confounding factors, the framework introduces temporal propensity matching, ensuring that the estimation of the causal effect between events is robust and less prone to spurious correlations. By fine-tuning large pre-trained language models to sample and analyze event sequences, ROCK provides a modular, zero-shot approach for commonsense causality reasoning without task-specific training.

\paragraph{Causal Prompting~\cite{zhang2024causal}.} This paper introduces a novel method called Causal Prompting, designed to reduce biases in Large Language Models (LLMs) by utilizing the front-door adjustment from causal inference. Traditional methods for mitigating bias often focus on model training or fine-tuning, but this paper shifts the approach by employing causal reasoning directly on the prompts, without accessing the model’s parameters or output logits. \textbf{In this framework, the Chain-of-Thought (CoT) generated by the LLM acts as a mediator between the input prompt ($X$) and the predicted output ($A$).} The causal effect between $X$ and $A$ is split into two components: the effect of $X$ on CoT ($R$) and the effect of CoT ($R$) on the final output ($A$). The approach leverages contrastive learning to align the encoder space with that of LLMs, improving the estimation of causal effects. The model’s performance is evaluated across seven natural language processing (NLP) datasets, and experimental results show that causal prompting significantly outperforms existing debiasing methods.

Specifically, Causal Prompting first addresses the causal relationship between the input ($X$) and the output ($A$) by breaking down this relationship into two causal paths: $X$ to CoT ($R$) and CoT to $A$. The process begins with generating multiple CoTs based on the input prompt $X$, which are then clustered to identify the representative CoTs that capture the core reasoning patterns. These CoTs serve as mediators between the input and the output. To estimate the causal effect between $R$ and $A$, a normalized weighted geometric mean approximation is employed, alongside contrastive learning and self-consistency prompting. This ensures that the encoder’s representation space aligns with that of the LLM, allowing for more accurate estimation of the causal effect and reducing bias in the model’s predictions.

\paragraph{CaCo-CoT~\cite{tang2023towards}.} The paper specifically addresses knowledge reasoning tasks, focusing on improving the reasoning capabilities of Large Language Models (LLMs) in tasks such as Science Question Answering (ScienceQA) and Commonsense Reasoning (Com2Sense) by promoting causal consistency through a multi-agent system. The proposed framework, called CaCo-CoT (Causal-Consistent Chain of Thought), involves multiple agents working collaboratively: reasoner agents that mimic human-like causality to solve problems, and an evaluator agent that assesses the logical consistency of the solutions by examining them from both causal and counterfactual perspectives. The reasoners break down complex problems into subquestions and generate answers through a structured reasoning process. The evaluator agent reviews these answers, challenging them with counterfactual scenarios to ensure the final answer is both accurate and causally consistent. The paper introduces a recursive evaluation process that relies on consensus among the agents. If a reasoner’s solution does not meet a predefined confidence threshold, the evaluator re-examines the solution for logical inconsistencies or contradictions in counterfactual reasoning. The final decision is made based on majority consensus and causal verification, helping LLMs avoid pitfalls such as inference errors and factual inaccuracies.

\paragraph{MAC~\cite{duong2024multi}.} The paper presents a novel framework, MAC (Multi-Agent Causality), for causal discovery by leveraging the multi-agent capabilities of Large Language Models (LLMs). MAC consists of three models: the Meta Agents Model, where agents debate to identify causal relationships; the Coding Agents Model, which selects and executes statistical algorithms for causal discovery; and the Hybrid Model, which combines both approaches. These models use expert reasoning and statistical methods to generate causal graphs. The key algorithm in the Meta Agents Model iterates through variables in a dataset, using agent debates to evaluate causal relationships, outputting a causal graph in matrix form. The Coding Agents Model involves two steps: selecting a causal discovery algorithm through agent debates and then executing it to generate the causal graph. The Hybrid Model integrates both methods, enhancing causal inference by combining reasoning with statistical execution.

\vspace{7pt}
In conclusion, while LLMs can effectively recite and retrieve causal knowledge embedded in their training data, their inherent lack of reasoning abilities limits their capacity for true causal inference. The need for thoughtfully crafted prompts highlights the human role in guiding LLMs to provide more meaningful and coherent responses. As research continues, the challenge remains to enhance the ability of LLMs to reason beyond spurious correlations and demonstrate a deeper understanding of causality. Further advancements in prompt engineering and model training strategies will be essential for LLMs to transition from mere information recall to authentic causal reasoning, enabling more robust and reliable applications of these models in real scenarios.

\section{Causality for Benchmark}
\label{Sec:Benchmark}
Causal reasoning is a critical task for understanding relationships between events, and evaluating the ability of large language models (LLMs) in this area has become an active field of research. Numerous benchmarks have been carefully developed to assess how effectively LLMs can identify, reason about, and infer causal relationships across diverse scenarios. In this section, we introduce recent, thoughtfully designed benchmarks used to evaluate LLMs’ causal reasoning capabilities, including their performance on tasks involving real-world events, structured datasets, and counterfactual reasoning.

%%%%%%%%%%%%%%%%%%%%%%%%%%%%%%%%%%%%%%%
%%%%%%%%%%%%%%%%%%%%%%%%%%%%%%%%%%%%%%%
%%%%%%%%%%%%%%%%%%%%%%%%%%%%%%%%%%%%%%%
{
\newcounter{table2}
\renewcommand{\thetable}{2.\arabic{table2}}

\setcounter{table2}{1}
\begin{table}[htbp]
\centering
\small
\begin{tabular}{|p{3cm}|p{3cm}|p{4cm}|p{5cm}|}
\hline
\textbf{Benchmark} & \textbf{LLM} & \textbf{Target} & \textbf{Tasks} \\ \hline
\href{https://github.com/epfl-nlp/CRAB}{\textcolor{blue}{Crab}} \cite{romanou2023crab} & Flan-Alpaca-GPT4-XL, GPT-3 (text-davinci-003), GPT-4 & Assessing the Strength of Causal Relationships Between Real-World Events & Pairwise Causal Discovery (yes/no); Pairwise Causal Strength (high/medium/low/ no); Pairwise Causal Score (0-100); Primary Causal Event (A/B/C/D) \\ \hline
\href{https://www.bnlearn.com/}{\textcolor{blue}{CausalBench}} \cite{zhou2024causalbench} & BERT, RoBERTa, DeBERTa, DistilBERT, LLAMA, OPT, InternLM, Falcon, GPT3.5-Turbo, GPT4 & Exploring the capabilities of LLMs in understanding causal relationships of varying depths and difficulties & Direct/Indirect Correlation (yes/no); Causal Discovery in Skeleton (yes/no); Causal Discovery in DAG (yes/no) \\ \hline
\href{https://huggingface.co/datasets/causalnlp/corr2cause}{\textcolor{blue}{CORR2CAUSE}} \cite{jin2024can} & BERT, RoBERTa, BART, DeBERTa, DistilBERT, DistilBART, GPT-3, GPT-3.5, GPT-4 & Evaluating the ability of large language models (LLMs) to infer causality from correlational statements & Pairwise Causal Hypothesis (yes/no) \\ \hline
\href{https://github.com/cicl-stanford/moca}{\textcolor{blue}{Moca}} \cite{nie2024moca} & RoBERTa, ALBERT, Electra, GPT-2, GPT-3, GPT-3.5, GPT-4, Alpaca-7B, Claude & Measuring Human-Language Model Alignment on Causal and Moral Judgment Tasks & Causal Judgment (yes/no); Moral Judgment (yes/no) \\ \hline
\href{https://github.com/causalNLP/cladder}{\textcolor{blue}{Cladder}} \cite{jin2024cladder} & GPT-4, GPT-3.5, GPT-3, LLaMa, Alpaca & Assessing causal reasoning in language models & Association (yes/no), Intervention (yes/no), Counterfactuals (yes/no) \\ \hline
\href{https://github.com/wyu97/IfQA}{\textcolor{blue}{IfQA}} \cite{yu2023ifqa} & Codex and ChatGPT & Assessing models’ abilities to handle counterfactual reasoning in open-domain question-answering (QA) tasks & Counterfactual Reasoning in IfQA (Entity/Number/ Date/Data) \\ \hline
\href{https://github.com/Eleanor-H/CLOMO}{\textcolor{blue}{Clomo}} \cite{huang2023clomo} & GPT-3.5-Turbo, GPT-4, GPT-4o, LLaMA, LLaMA2, Flan-T5, ChatGLM2, Baichuan2, InternLM, Vicuna-v1.5, Qwen, WizardLM & Assessing the counterfactual reasoning capabilities of large language models (LLMs) & Counterfactual Reasoning in Modification (Necessary Assumption/Sufficient Assumption/Strength/Weaken) \\ 
\hline
\href{https://github.com/apergo-ai/CRASS-data-set}{\textcolor{blue}{Crass}} \cite{frohberg2022crass} & BART, RoBERTa, MPNet, DeBERTa v3, GPT-3, T0pp & Assessing the counterfactual reasoning capabilities of large language models (LLMs) & Counterfactual Reasoning (choose correct answer) \\ \hline
\href{https://github.com/xxxiaol/QRData}{\textcolor{blue}{QRData}} \cite{liu2024llms} & GPT-4, GPT-3.5 Turbo, Gemini-Pro, Llama-2-chat, WizardMath, Deepseek-coder-instruct, CodeLlama-instruct, TableLlama, and AgentLM & Evaluating the statistical and causal reasoning abilities of large language models (LLMs) & Statistical Reasoning (Probability/Distribution/Es- timation/Hypothesis Testing/ Prediction); Causal Reasoning (Confounding/Causal Discovery / Causal Effect Estimation / Instrumental Variables / Panel Data); Causal Effect Estimation (ATE/ATT/ATC) \\ \hline
\end{tabular}
\caption{Overview of Causal Reasoning Benchmarks for LLMs}
\label{tab:benchmark1}
\end{table}

\stepcounter{table2}
\begin{table}[htbp]
\centering
\small
\begin{tabular}{|p{3cm}|p{3cm}|p{4cm}|p{5cm}|}
\hline
\textbf{Benchmark} & \textbf{LLM} & \textbf{Target} & \textbf{Tasks} \\ \hline
\href{https://github.com/ArrogantL/ChatGPT4CausalReasoning}{\textcolor{blue}{CausalReasoning}} \cite{gao2023chatgpt} & BERT, RoBERTa, LLaMA, FLAN-T5, GPT-2, GPT-3,GPT-3.5, GPT-4 & Evaluating the causal reasoning capabilities of ChatGPT & Event Causality Identification (yes/no); Causal Discovery (Multiple Choice/Binary Classification); Causal Explanation Generation (Text Generation) \\ \hline
\href{https://huggingface.co/datasets/voidful/StrategyQA}{\textcolor{blue}{Counterfactual Simulatability}} \cite{chen2023models} \newline \href{https://github.com/kawine/dataset_difficulty}{} & GPT-3.5, GPT-4 & Evaluating the counterfactual simulatability of natural language explanations & Multi-hop Factual Reasoning (Text Generation); Reward Modeling (Text Generation) \\ \hline
\href{https://github.com/py-why/pywhy-llm}{\textcolor{blue}{New Frontier}} \cite{kiciman2023causal} & GPT-3, GPT-3.5, GPT-4 & Opening a new frontier for causality & Pairwise Causal Discovery (yes/no or A/B); Counterfactual Reasoning (choose correct answer) \\ \hline
\href{https://github.com/MoritzWillig/causalParrots/}{\textcolor{blue}{Causal Parrots}} \cite{zevcevic2023causal} & Luminous, OPT, GPT-3, GPT-4 & Investigating if large language models (LLMs) truly understand causality or just mimic learned correlations & Common Sense Inference (yes/no); Causal Discovery (causal/non-causal); Knowledge Base Fact Embeddings (causal/anti-causal) \\ \hline
\href{https://opencausalab.github.io/CaLM}{\textcolor{blue}{CaLM}} \cite{chen2024causal} & Baichuan1, Baichuan2, InternLM, Llama 2, Qwen, Koala, Wizardcoder, Vicuna, GPT-3, GPT-3.5-Turbo, GPT-4, Claude2 & Evaluating causal reasoning capabilities of large language models (LLMs) & Four Question Types in 92 Causal Targets: Binary classification, Choice selection, Open-ended generation, and Probability calculation \\ \hline
\href{https://github.com/zharry29/causal_reasoning_of_entities_and_events}{\textcolor{blue}{CREPE}} \cite{zhang2023causal} & T5, T0, GPT-3, ChatGPT, Codex & Assessing the ability of large language models (LLMs) to reason causally about events and entities in procedural texts & Causal Reasoning in Procedural Texts (More Likely/Less Likely/Equally Likely); Factual Reasoning in Procedural Texts (True/False or Yes/No) \\ \hline
\href{https://github.com/ncsulsj/Causal_LLM}{\textcolor{blue}{Knowledge}} \cite{cai2023knowledge} & GPT-3.5, GPT-4, GPT-4 Turbo, Claude 2, LLaMa2, Mistral & Exploring the causal reasoning of large language models (LLMs) & Causal Discovery (Omit Knowledge, Omit Data); Pairwise Causal Discovery (Causal Direction); Reverse Causal Discovery (Causal Direction) \\ \hline
\href{https://github.com/junwang4/causal-language-use-in-science}{\textcolor{blue}{Causal Language}} \cite{kim2023can} \newline \href{https://github.com/junwang4/correlation-to-causation-exaggeration}{} & GPT-3.5, ChatGPT & Evaluating ChatGPT’s ability to understand causal language in science papers and news & Understand Causal Language (direct causal/conditional causal/correlational/no relationship) \\ \hline
\end{tabular}
\caption{Overview of Causal Reasoning Benchmarks for LLMs}
\label{tab:benchmark2}
\end{table}
}
%%%%%%%%%%%%%%%%%%%%%%%%%%%%%%%%%%%%%%%
%%%%%%%%%%%%%%%%%%%%%%%%%%%%%%%%%%%%%%%
%%%%%%%%%%%%%%%%%%%%%%%%%%%%%%%%%%%%%%%

In Tables \ref{tab:benchmark1}-\ref{tab:benchmark2}, we provide a high-level overview of the benchmarks that focus on causal reasoning, including the datasets, models used, and tasks they assess. This table offers a condensed summary of each benchmark, with further details provided in the subsequent enumerated list. As shown in the Tables \ref{tab:benchmark1}-\ref{tab:benchmark2}, these benchmarks cover a broad range of causal reasoning tasks. The table gives a quick snapshot of the benchmarks, but more detailed explanations of each benchmark are provided in the following list:

\begin{enumerate}
    \item \textbf{Assessing the Strength of Causal Relationships Between Real-World Events} \cite{romanou2023crab}: CRAB (\url{https://github.com/epfl-nlp/CRAB}) consists of 173 documents covering 20 different real-world newsworthy events. It includes 384 unique extracted event instances and 2,730 pairwise causality scores between these events. For each event pair, causality scores provided by multiple annotators were averaged to create a unique scalar judgment. Based on this score, each event pair is categorized into one of four causality classes: high, medium, low, or no causality. They find that state-of-the-art language models perform poorly in pairwise causal inference and responsibility assignment when events are spread across documents. Furthermore, this weak performance is amplified when LLMs must identify causal relationships in complex causal structures rather than simple linear chains.
    \item \textbf{Exploring the capabilities of LLMs in understanding causal relationships of varying depths and difficulties} \cite{zhou2024causalbench}: CausalBench (\url{https://www.bnlearn.com/}) studies 15 real-world causal discovery datasets, ranging from 2 to 109 nodes, including benchmarks like Asia, Cancer, and Pathfinder. Each dataset includes structured data and domain-specific background knowledge for evaluating model performance. With over 500,000 data points, CausalBench challenges models to identify correlations, reconstruct causal skeletons, and determine causal directions between variables. LLMs for causal learning fall short of human performance. Closed-source LLMs significantly surpass open-source ones but still fall short of the current performance of the classic and state-of-the-art (SOTA) causal learning methods.
    \item \textbf{Evaluating the ability of large language models (LLMs) to infer causality from correlational statements} \cite{jin2024can}: CORR2CAUSE (\url{https://huggingface.co/datasets/causalnlp/corr2cause}) consists of 207,972 samples focused on inferring causality from correlational data. Each sample includes correlational statements between variables and a hypothesis about their causal relationship. The dataset is generated using directed graphical causal models (DGCMs) with 2 to 6 variables, testing six causal relations: Is-Parent, Is-Ancestor, Is-Child, Is-Descendant, Has-Collider, and Has-Confounder. Large language models (LLMs) perform poorly in inferring causality from correlational data, often achieving results close to random guessing, especially in out-of-distribution scenarios. Fine-tuning LLMs improves their performance on the CORR2CAUSE dataset, but these improvements are not robust, and the models fail to generalize causal reasoning beyond the specific patterns present in their training data.
    \item \textbf{Measuring Human-Language Model Alignment on Causal and Moral Judgment Tasks} \cite{nie2024moca}: MoCa (\url{https://github.com/cicl-stanford/moca}) consists of 144 causal judgment stories and 62 moral judgment stories, totaling 206 stories. Each story is annotated by experts for key factors influencing causal and moral judgments, such as norm type, event normality, and action or omission. Then, 25 yes/no responses were collected from participants for each story, resulting in a total of 5,150 human responses in the dataset. While the models showed improved alignment overall, the analysis revealed significant differences in how models weigh different factors compared to humans, indicating room for improvement in causal and moral reasoning tasks.
    \item \textbf{Assessing causal reasoning in language models} \cite{jin2024cladder}: CLADDER (\url{https://github.com/causalNLP/cladder}) consists of 10,112 binary causal questions across three levels of Pearl’s Ladder of Causation: associational, interventional, and counterfactual. Generated from over 50 causal graphs, the dataset covers common structures such as confounding, mediation, and colliders. Each sample includes a causal query, a ground-truth answer derived through a causal inference engine, and step-by-step reasoning explanations. While large language models (LLMs) have made significant advancements, they still struggle with formal causal reasoning, particularly in more complex areas like counterfactuals. Although CAUSALCOT improves performance by guiding the models through structured reasoning, the results underscore that LLMs remain inadequate for reliable, high-level causal reasoning.
    \item \textbf{Assessing models’ abilities to handle counterfactual reasoning in open-domain question-answering (QA) tasks} \cite{yu2023ifqa}: IfQA (\url{https://github.com/wyu97/IfQA}) consists of 3,800 questions based on counterfactual presuppositions, extracted from relevant Wikipedia passages. Each question is framed using an “if” clause, requiring models to go beyond factual retrieval to perform reasoning based on hypothetical situations. The dataset is divided into two splits: a supervised setting (IfQA-S) and a few-shot setting (IfQA-F). Existing LLMs, including ChatGPT, face significant challenges in both retrieval and reasoning, especially when it comes to handling counterfactual reasoning in IfQA. Although the chain-of-thought approach improves large language models’ ability to reason through counterfactual scenarios, closed-book models still struggle due to their lack of access to external knowledge. Passage retriever (BM25) + Large model reasoner performs the best on IfQA.
    \item \textbf{Assessing the counterfactual reasoning capabilities of large language models (LLMs)} \cite{huang2023clomo}: CLoMo (\url{https://github.com/Eleanor-H/CLOMO}) contains 1,000 manually constructed high-quality data points, each containing argumentative text with associated premises. For each data point, models are tasked with modifying the text to satisfy a specific logical relationship between two premises. The dataset includes four main logical relations: necessary assumption, sufficient assumption, strengthening, and weakening. Large Language Models (LLMs) demonstrate some capability in counterfactual reasoning tasks, such as the CLOMO task, with GPT-4 outperforming human performance in certain logical relations. However, there is still a significant gap in handling more complex logical reasoning, particularly in challenging relations like “Sufficient Assumption” and “Weaken.” Fine-tuning models can significantly improve their performance, but further development and more high-quality counterfactual reasoning datasets are needed to better enhance the counterfactual logical reasoning abilities of current models.
    \item \textbf{Assessing the counterfactual reasoning capabilities of large language models (LLMs)} \cite{frohberg2022crass}: CRASS (\url{https://github.com/apergo-ai/CRASS-data-set}) consists of 274 premise-counterfactual tuples (PCTs), each containing a base premise, a counterfactual antecedent, and three possible consequents: one correct and two distractors. The dataset challenges models to predict the correct consequent based on hypothetical changes to real-world events. Existing LLMs, such as GPT-3 and T0pp, exhibit significant deficiencies in addressing counterfactual reasoning tasks, with a performance gap of over 25\% compared to human baselines. While some models perform relatively better in dynamic and impossible scenarios, overall, there remains substantial room for improvement in LLMs’ understanding and reasoning of complex causal relationships, with the CRASS benchmark providing a valuable tool for guiding future model development.
    \item \textbf{Evaluating the statistical and causal reasoning abilities of large language models (LLMs)} \cite{liu2024llms}: QRData (\url{https://github.com/xxxiaol/QRData}) consists of 411 questions across 195 real-world data sheets, covering both numerical and multiple-choice questions, with 142 focusing on statistical tasks and 269 on causal inference. Each question is paired with a data sheet from textbooks, online resources, or academic papers. QRData is accompanied by an auxiliary set, QRTEXT, which contains 290 text-based reasoning questions. Large language models (LLMs) models have difficulties in data analysis and causal reasoning. Although GPT-4 outperforms open-source models, achieving up to 58\% accuracy, there remains much room for improvement, especially in data analysis and causal reasoning. Even though powerful LLMs like GPT-4 have acquired causal knowledge, they can hardly integrate them with the provided data.
    \item \textbf{Evaluating the causal reasoning capabilities of ChatGPT} \cite{gao2023chatgpt}: This paper (\url{https://github.com/ArrogantL/ChatGPT4CausalReasoning}) studies 5 widely-used causal discovery datasets: EventStoryLine contains 22 topics, 258 documents, 5,334 events, and 1,770 causal event pairs; Causal-TimeBank contains 184 documents, 6,813 events, and 318 causal event pairs; MAVEN-ERE contains 90 topics, 4,480 documents, 103,193 events, and 57,992 causal event pairs; COPA consists of 1,000 multiple-choice questions centered on everyday life scenarios; and e-CARE contains 21,324 multiple-choice questions across various domains, designed for evaluating causal explanation generation. ChatGPT is not a good causal reasoner, but is good at causal explanation generation; ChatGPT has a serious causal hallucination, possibly due to the causal reporting biases and ChatGPT’s upgrading processes. The ICL and CoT techniques can further exacerbate such causal hallucination. For events in sentences, ChatGPT excels at capturing explicit causality and performs better in sentences with lower event density and smaller event distances.
    \item \textbf{Evaluating the counterfactual simulatability of natural language explanations} \cite{chen2023models}: This paper (\url{https://huggingface.co/datasets/voidful/StrategyQA}) studies 2 datasets for evaluating LLMs’ counterfactual reasoning. StrategyQA consists of 2,780 yes/no questions requiring implicit multi-step reasoning, with annotated reasoning steps and supporting evidence. Stanford Human Preference (SHP) includes 28,000 examples of human preferences for agent responses, used to assess reward models. Both datasets test models’ abilities to align with human reasoning and predict behavior on counterfactuals. State-of-the-art LLMs generate misleading explanations that lead to wrong mental models, and thus there is plenty of room for improvement for our metrics. Counterfactual simulatability does not correlate with plausibility, and thus RLHF methods that satisfy humans may not improve counterfactual simulatability.
    \item \textbf{Opening a new frontier for causality} \cite{kiciman2023causal}: This paper (\url{https://github.com/py-why/pywhy-llm}) evaluates LLMs using 4 datasets. The Tübingen cause-effect pairs dataset includes 108 cause-effect pairs from 37 datasets across various domains, testing causal direction inference. The Neuropathic pain dataset has 221 variables and 475 expert-annotated causal relationships between nerve conditions and symptoms. The Arctic sea ice dataset includes 12 variables and 48 edges related to factors influencing Arctic sea ice thickness. The CRASS dataset consists of 274 premise-counterfactual tuples, each with a base premise, a counterfactual antecedent, and three possible consequents. LLMs show strong potential in causal reasoning, outperforming traditional methods in some tasks, but they still have limitations like logical errors and inconsistencies. They can assist in generating causal graphs and handling specific tasks but are not yet fully reliable for complex causal reasoning.
    \item \textbf{Investigating if large language models (LLMs) truly understand causality or just mimic learned correlations} \cite{zevcevic2023causal}: This paper (\url{https://github.com/MoritzWillig/causalParrots/}) uses a dataset of 20 questions structured to evaluate causal reasoning for Common Sense Inference Tasks, with variable chains ranging from 3 to 10 in length. For Causal Discovery on Ground Truth, the paper employs six datasets—Altitude, Health, Driving, Recovery, Cancer, and Earthquake—each containing between 10 and 100 variables with known causal relationships. For Knowledge Base Fact Embeddings, the paper utilizes 16,567 cause-effect pairs from ConceptNet, embedded into five sentence templates, resulting in 165,670 causal and anti-causal samples. Large language models (LLMs) can mimic causal relationships but do not genuinely understand or perform causal reasoning, often relying on learned correlations from training data rather than true causal inference. They might be “causal parrots” but rather underwhelming ones in that they do not recite everything that one would want them to.
    \item \textbf{Evaluating causal reasoning capabilities of large language models (LLMs)} \cite{chen2024causal}: CaLM (\url{https://opencausalab.github.io/CaLM}) consists of 126,334 data samples spanning 92 causal targets across 46 causal tasks. These tasks cover three distinct text modes—Natural, Symbolic, and Mathematical—and involve two languages: English and Chinese. Each causal target is associated with multiple adaptations and metrics, providing a comprehensive framework for evaluating causal reasoning capabilities. While LLMs exhibit substantial capabilities in causal reasoning, there are notable limitations in their performance across different tasks and settings. LLMs perform better on simpler, more structured tasks but struggle with complex causal reasoning, especially in counterfactual and intervention-based scenarios. The models show variability in handling different languages and modes, such as natural, symbolic, and mathematical texts. Additionally, scale and adaptation techniques significantly influence model accuracy, with larger models and specialized prompting yielding better results.
    \item \textbf{Assessing the ability of large language models (LLMs) to reason causally about events and entities in procedural texts} \cite{zhang2023causal}: CREPE (\url{https://github.com/zharry29/causal_reasoning_of_entities_and_events}) consists of 183 procedural texts across various domains, including 1,219 individual steps and 324 event likelihood changes paired with corresponding entity state changes. The dataset spans diverse topics such as recipes, household activities, and technology, each procedure containing an average of 6-7 steps. Event likelihoods are categorized into three labels: more likely, less likely, or equally likely, based on changes in the entity states. Large language models (LLMs), including GPT-3 and Codex, struggle with causal reasoning in procedural texts, particularly in predicting event likelihoods based on entity state changes. While LLMs perform better when provided with structured, code-like prompts that represent procedural steps, their performance still lags significantly behind human reasoning.
    \item \textbf{Exploring the causal reasoning of large language models (LLMs)} \cite{cai2023knowledge}: This paper (\url{https://github.com/ncsulsj/Causal_LLM}) studies 9 datasets from various domains, including Galton with 898 samples on family height measurements (4 nodes, 3 causal relations), Sachs with 7,466 biological samples on phosphorylated proteins (12 nodes, 20 relations), Alcohol with 345 health samples related to alcohol consumption (6 nodes, 5 relations), EcoSystem with 721 environmental samples on carbon flux (4 nodes, 3 relations), MPG with 392 engineering samples on vehicle attributes and fuel consumption (5 nodes, 6 relations), DWD with 349 geographical samples on climate factors (6 nodes, 6 relations), Cement with 1,030 engineering samples on material strength (9 nodes, 8 relations), Stock with 1,331 financial samples on stock returns (5 nodes, 3 relations), and Arrhythmia with 450 biological samples on cardiac health (4 nodes, 3 relations). LLMs’ causal reasoning ability mainly depends on the context and domain-specific knowledge provided. In the absence of such knowledge, LLMs can still maintain a degree of causal reasoning using the available numerical data, albeit with limitations in the calculations. This motivates the proposed fine-tuned LLM for pairwise causal discovery, effectively leveraging both knowledge and numerical information.
    \item \textbf{Evaluating ChatGPT’s ability to understand causal language in science papers and news} \cite{kim2023can}: This paper (\url{https://github.com/junwang4/causal-language-use-in-science}) studies 2 open-access cross-genre datasets: a sample of 3,061 sentences from PubMed abstracts and 2,076 sentences from health-related press releases on EurekAlert. Both datasets are manually annotated with four categories: direct causal, conditional causal, correlational, and no relationship. ChatGPT has a promising but still limited ability to understand causal language in science writing. CoTs improved prompt performance, but finding the optimal prompt is difficult with inconsistent results and the lack of effective methods to establish cause-effect between prompts and outcomes.
\end{enumerate}

In summary, these benchmarks provide diverse and complex tasks to evaluate how well LLMs can handle causal reasoning across different contexts, from real-world events to counterfactual and correlational data. Each benchmark highlights different strengths and weaknesses of LLMs in terms of their reasoning capabilities, paving the way for future improvements in the field.

\section{Future Directions on LLMs with Causality}
\label{sec:future}

In previous sections, we have shown that incorporating causality into LLMs across five key stages — pretraining, fine-tuning, alignment, inference, and evaluation — can help alleviate these challenges. By embedding causal reasoning capabilities, models can gain a more structured understanding of relationships between variables, leading to more reliable outputs and reducing the likelihood of hallucinations. This structured approach would allow LLMs to handle complex contexts better, improve generalization, and create safer, more trustworthy systems, particularly when applied in sensitive domains where accountability is critical. Thus, in this section, we provide six promising future directions to further enhance its causal reasoning capabilities.

\paragraph{Developing Causal Transformer for Enhancing Foundation Model}
In Section \ref{sec:CFM}, we have introduced two frameworks that integrate causal reasoning into transformer-based models~\cite{rohekar2024causal, zhang2024towards}. Rohekar et al \cite{rohekar2024causal} found the attention matrix $A$ is analogous to the causal dependencies encoded in $ (I - G)^{-1} \Lambda $, making self-attention a mechanism that captures causal relationships between input tokens. Meanwhile, \cite{zhang2024towards} introduced the duality between optimal balancing and attention. While they establish a connection between causality and the attention module, they still rely on the attention mechanism to capture the interactions between tokens. This reliance can lead to the model inheriting demographic biases and social stereotypes present in the training data, limiting its effectiveness in providing unbiased and accurate predictions. To overcome these challenges, a promising direction is to move from a purely correlational framework to one that prioritizes causal relationships. By focusing on causal links, we can mitigate the effects of biases and enhance the model’s ability to identify meaningful interactions among tokens. One effective approach is to leverage the Attention-Based Causal Discovery (ABCD) algorithm~\cite{rohekar2024causal}, which imposes constraints on the attention mechanism. These constraints guide the model to learn representations that align more closely with causal matrices. By embedding these constraints, attention scores can be made to reflect true causal relationships, not mere correlations. This shift not only helps in reducing biases but also improves the model’s interpretability and reliability, leading to more accurate and equitable predictions in a variety of applications.

\paragraph{Multiple Task Learning with Causal Graph} 
On the internet, we have accumulated a vast amount of knowledge graphs and causal knowledge data. Embedding this knowledge regarding causal structures into the representation space ensures that the model’s internal representations accurately reflect the underlying causal relationships, leading to more informed decision-making. To enhance large language models (LLMs) with a deeper understanding of causality, multiple-task learning (MTL) frameworks can be employed. MTL allows the model to learn from various related tasks simultaneously, improving generalization and task-specific performance. For instance, Zhao et al.~\cite{zhao2017constructing}  introduced a method for constructing and embedding abstract event causality networks from textual data. Their approach generalizes specific events into abstract patterns using a causality network, which is then embedded into a continuous vector space. This method allows models to learn high-level, frequent causality patterns that enhance their ability to detect and predict causal relationships. By embedding these abstract causality networks into LLMs through MTL, the model can capture both specific and generalized causal patterns, leading to more robust and interpretable predictions.

\paragraph{Reverse Causal Relations for Address Reversal Curse}
Please find the Reversal Curse example in Table \ref{tab:issues}. Due to the autoregressive nature of large language models (LLMs), these models can predict the next token or line with high accuracy. For instance, an LLM can tell us that the line following Line A in the US anthem is Line B. However, it struggles to identify that the line before Line B is Line A. This phenomenon, known as the reversal curse~\cite{berglund2023reversal}, stems from the model’s left-to-right training, which limits its ability to reverse sequences and predict information backward. While this may go unnoticed for frequently repeated facts, it poses a significant challenge for less common knowledge, which may only appear in one direction. To address this limitation, researchers at Meta proposed several approaches, such as reversing sequences through word reversal, entity reversal, and random reversal, to train models in the opposite direction~\cite{golovneva2024reverse}. These methods provide a way for models to learn bidirectionally, potentially overcoming the reversal curse.

However, while reversing sequences shows promise, there is a more nuanced approach to be explored: reversing sentences or entities based on event relationships, temporal order, and causality. Instead of purely mechanical reversals, incorporating logical, causality-based reversals could offer deeper improvements in LLM performance. For instance, reversing relations such as cause-effect into effect-cause could introduce counterfactual data, allowing models to better understand the nature of events and predict outcomes in both directions. Similarly, reversing relations such as entity-destination or content-container could improve models’ reasoning capabilities in tasks that involve knowledge retrieval or reasoning across sequential steps. By intelligently applying reversals to causal and event-driven relationships, we can craft a more logical and balanced corpus for training LLMs, helping them overcome the inherent biases of autoregressive training. Here, reasearchs summarized some candidate relations: message-topic, topic-message, entity-destination, destination-entity, content-container, container-content, effect-cause, cause-effect, whole-component, component-whole, collection-member, member-collection, agency-instrument, instrument-agency, producer-product, product-producer, entity-origin, origin-entity. We might consider reversing these relations to create a more logical and reasonable counterfactual corpus. Using these reversals, we can develop a corpus that better addresses autoregression issues and improves model performance.

\paragraph{Instrumental Variable Learning for RLHF}
In the alignment stage of large language models (LLMs), Reinforcement Learning from Human Feedback (RLHF) is commonly used to refine the model’s behavior based on human input. RL concepts have strong connections with causal inference, where the agent’s actions are akin to treatments, the environment acts as a confounder, and cumulative rewards represent the outcomes. Applying causal reinforcement learning (Causal RL) algorithms to RLHF can enhance the alignment process by addressing underlying causal structures and reducing biases introduced by confounders.

When the assumption of unconfoundedness holds, various methods can estimate the state-action reward reliably. Without unconfoundedness assumption, instrumental variable learning still can be leveraged for policy optimization. IV techniques allow for the estimation of causal effects in the presence of unmeasured confounders by introducing instruments—variables that affect the agent’s action but are independent of the confounders. For example, recent approaches have utilized IVs in offline policy evaluation (OPE), improving Q-function estimators~\cite{chen2021instrumental} and leading to the development of algorithms such as the IV-aided Value Iteration (IVVI) algorithm~\cite{liao2021instrumental}. These advancements offer promising directions for RLHF by correcting bias and optimizing policies in the face of time-dependent noise and confounding factors. In the future, integrating IV techniques into RLHF can provide LLMs with better alignment by refining feedback-driven learning, ensuring more reliable and accurate human-aligned responses.

\paragraph{Using Prompt as Instruments or Mediators} In the future, prompts could serve as powerful tools for understanding and applying causal inference in large language models (LLMs). Prompts are often treated as static inputs, but their role as variables influencing the decision-making process of LLMs has been largely overlooked. Different prompts can lead to varied responses from LLMs, significantly affecting the final outcome. By studying how different prompts impact responses, we can treat prompts as instrumental variables (IVs), particularly when analyzing human behavioral data. These prompts can serve as instruments that help uncover causal relationships between input and output by identifying variables that influence decisions while being independent of confounders.

Similarly, as demonstrated by Causal Prompting~\cite{zhang2024causal}, prompts can also act as mediators in the decision-making process. When viewed as part of a task’s intermediate steps, prompts bridge the gap between the input (X) and the final output (A), much like mediators in causal inference. By analyzing the causal effect of the prompt on the model’s reasoning process (such as Chain-of-Thought generation) and its eventual impact on the output, we can better estimate causal relationships and reduce biases in predictions. This dual role of prompts—as either instruments or mediators—opens new directions for improving LLM alignment, debiasing, and task performance by leveraging causal graphs and inference techniques.

\paragraph{Studying Self-Correction Capabilities of Large Language Models} Li et al \cite{li2024confidence} explores the intrinsic self-correction capabilities of large language models (LLMs) and proposes a confidence-based framework called If-or-Else (IoE) prompting. The core idea is that LLMs can assess the confidence in their own answers and, based on this self-assessment, decide whether to retain or revise their responses. This method addresses the issue of over-correction caused by excessive self-doubt, making the self-correction process more reliable. By incorporating this confidence-based mechanism, LLMs show significant improvements in accuracy, particularly in scenarios with low confidence, and excel in multi-task learning. The Self-Correction technique has a natural connection with causal reasoning. Confidence can be viewed as a latent variable that influences whether the model decides to revise its answer during self-correction. This is analogous to latent variable models in causal inference, where unobserved factors are adjusted to improve the accuracy and robustness of decisions. In future research, confidence could be explored as a latent variable within causal frameworks, combined with LLMs’ self-correction capabilities to better understand and model how causal factors impact decision-making processes.

\vspace{7pt}
Despite the significant progress in large language models (LLMs), several challenges persist, particularly concerning bias, decision-making transparency, and effective alignment with human feedback. Issues like the reversal curse, where LLMs struggle with backward reasoning due to their autoregressive nature, and biases inherited from training data, underscore the need for more robust solutions. To push LLMs closer to artificial general intelligence (AGI) and a more human-like intelligence, we outline six future directions focused on integrating causal reasoning. By leveraging causal inference techniques, these approaches aim to improve LLMs’ ability to understand complex relationships, make more informed decisions, and self-correct effectively. The proposed strategies will enable models to better emulate human cognitive processes, moving beyond simple correlations and towards understanding causal links that drive reasoning.

\section{Conclusion}

In this paper, we present a comprehensive review of how incorporating causality can enhance large language models (LLMs) at each stage of their lifecycle—from token embedding and foundational model training to fine-tuning, alignment, inference, and evaluation. We focus on several key areas: the use of debiased token embeddings and counterfactual training corpora during the pre-training phase to mitigate biases and improve causal feature learning; fine-tuning techniques such as Causal Effect Tuning (CET) and Distilling Counterfactuals (DISCO), which preserve foundational knowledge while adapting models to domain-specific tasks that require more profound causal reasoning; and alignment strategies, like Causal Preference Optimization (CPO), which employ causal inference to align ethical considerations with user preferences. Furthermore, we discuss the application of causal discovery methods to enhance inference by distinguishing between correlational and causal relationships, and the integration of counterfactual reasoning to facilitate more reflective and adaptable decision-making processes. Finally, we propose six promising future directions to further advance the causal reasoning capabilities of LLMs.

The integration of causal reasoning into LLMs represents a paradigm shift, enabling models to transcend mere statistical correlations and engage in structured, causal reasoning. While traditional models such as ChatGPT, LLaMA, PaLM, Claude, and Qwen are highly effective in language understanding and generation through their ability to recognize token-level patterns in large datasets, they often falter in tasks requiring profound causal comprehension. These models struggle with discerning the underlying causal relationships that are critical in domains like policy analysis, scientific research, and healthcare. By embedding causal reasoning, LLMs can deliver more reliable and contextually meaningful outputs, particularly in high-stakes fields where accurate causal understanding is paramount.

Thus, integrating causality into LLMs marks a significant frontier in AI research, enabling these models to reason about causal relationships and produce outputs that are not only more accurate but also contextually appropriate and robust. Infusing causal knowledge throughout the model lifecycle—from pre-training and fine-tuning to inference and alignment—allows LLMs to go beyond pattern recognition and address the complexities of real-world problems with a deeper level of reasoning. This causality-driven approach unlocks new potentials for LLMs to make significant contributions to critical areas such as healthcare, scientific discovery, and policy-making, where the ability to discern causal relationships is essential for informed decision-making.

{
\bibliography{reference.bib}

\begin{thebibliography}{100}
\providecommand{\url}[1]{#1}
\csname url@samestyle\endcsname
\providecommand{\newblock}{\relax}
\providecommand{\bibinfo}[2]{#2}
\providecommand{\BIBentrySTDinterwordspacing}{\spaceskip=0pt\relax}
\providecommand{\BIBentryALTinterwordstretchfactor}{4}
\providecommand{\BIBentryALTinterwordspacing}{\spaceskip=\fontdimen2\font plus
\BIBentryALTinterwordstretchfactor\fontdimen3\font minus \fontdimen4\font\relax}
\providecommand{\BIBforeignlanguage}[2]{{%
\expandafter\ifx\csname l@#1\endcsname\relax
\typeout{** WARNING: IEEEtran.bst: No hyphenation pattern has been}%
\typeout{** loaded for the language `#1'. Using the pattern for}%
\typeout{** the default language instead.}%
\else
\language=\csname l@#1\endcsname
\fi
#2}}
\providecommand{\BIBdecl}{\relax}
\BIBdecl

\bibitem{le2023bloom}
T.~Le~Scao, A.~Fan, C.~Akiki, E.~Pavlick, S.~Ili{\'c}, D.~Hesslow, R.~Castagn{\'e}, A.~S. Luccioni, F.~Yvon, M.~Gall{\'e} \emph{et~al.}, ``Bloom: A 176b-parameter open-access multilingual language model,'' 2023.

\bibitem{touvron2023llama}
H.~Touvron, T.~Lavril, G.~Izacard, X.~Martinet, M.-A. Lachaux, T.~Lacroix, B.~Rozi{\`e}re, N.~Goyal, E.~Hambro, F.~Azhar \emph{et~al.}, ``Llama: Open and efficient foundation language models,'' \emph{arXiv preprint arXiv:2302.13971}, 2023.

\bibitem{touvron2023llama2}
H.~Touvron, L.~Martin, K.~Stone, P.~Albert, A.~Almahairi, Y.~Babaei, N.~Bashlykov, S.~Batra, P.~Bhargava, S.~Bhosale \emph{et~al.}, ``Llama 2: Open foundation and fine-tuned chat models,'' \emph{arXiv preprint arXiv:2307.09288}, 2023.

\bibitem{zhao2023survey}
W.~X. Zhao, K.~Zhou, J.~Li, T.~Tang, X.~Wang, Y.~Hou, Y.~Min, B.~Zhang, J.~Zhang, Z.~Dong \emph{et~al.}, ``A survey of large language models,'' \emph{arXiv preprint arXiv:2303.18223}, 2023.

\bibitem{huang2023survey}
L.~Huang, W.~Yu, W.~Ma, W.~Zhong, Z.~Feng, H.~Wang, Q.~Chen, W.~Peng, X.~Feng, B.~Qin \emph{et~al.}, ``A survey on hallucination in large language models: Principles, taxonomy, challenges, and open questions,'' \emph{arXiv preprint arXiv:2311.05232}, 2023.

\bibitem{chang2024survey}
Y.~Chang, X.~Wang, J.~Wang, Y.~Wu, L.~Yang, K.~Zhu, H.~Chen, X.~Yi, C.~Wang, Y.~Wang \emph{et~al.}, ``A survey on evaluation of large language models,'' \emph{ACM Transactions on Intelligent Systems and Technology}, vol.~15, no.~3, pp. 1--45, 2024.

\bibitem{vaswani2017attention}
A.~Vaswani, ``Attention is all you need,'' \emph{Advances in Neural Information Processing Systems}, 2017.

\bibitem{devlin2018bert}
J.~Devlin, ``Bert: Pre-training of deep bidirectional transformers for language understanding,'' \emph{arXiv preprint arXiv:1810.04805}, 2018.

\bibitem{raffel2020exploring}
C.~Raffel, N.~Shazeer, A.~Roberts, K.~Lee, S.~Narang, M.~Matena, Y.~Zhou, W.~Li, and P.~J. Liu, ``Exploring the limits of transfer learning with a unified text-to-text transformer,'' \emph{Journal of machine learning research}, vol.~21, no. 140, pp. 1--67, 2020.

\bibitem{radford2019language}
A.~Radford, J.~Wu, R.~Child, D.~Luan, D.~Amodei, I.~Sutskever \emph{et~al.}, ``Language models are unsupervised multitask learners,'' \emph{OpenAI blog}, vol.~1, no.~8, p.~9, 2019.

\bibitem{brown2020language}
T.~B. Brown, ``Language models are few-shot learners,'' \emph{arXiv preprint arXiv:2005.14165}, 2020.

\bibitem{achiam2023gpt}
J.~Achiam, S.~Adler, S.~Agarwal, L.~Ahmad, I.~Akkaya, F.~L. Aleman, D.~Almeida, J.~Altenschmidt, S.~Altman, S.~Anadkat \emph{et~al.}, ``Gpt-4 technical report,'' \emph{arXiv preprint arXiv:2303.08774}, 2023.

\bibitem{chowdhery2023palm}
A.~Chowdhery, S.~Narang, J.~Devlin, M.~Bosma, G.~Mishra, A.~Roberts, P.~Barham, H.~W. Chung, C.~Sutton, S.~Gehrmann \emph{et~al.}, ``Palm: Scaling language modeling with pathways,'' \emph{Journal of Machine Learning Research}, vol.~24, no. 240, pp. 1--113, 2023.

\bibitem{bai2023qwen}
J.~Bai, S.~Bai, Y.~Chu, Z.~Cui, K.~Dang, X.~Deng, Y.~Fan, W.~Ge, Y.~Han, F.~Huang \emph{et~al.}, ``Qwen technical report,'' \emph{arXiv preprint arXiv:2309.16609}, 2023.

\bibitem{minaee2024large}
S.~Minaee, T.~Mikolov, N.~Nikzad, M.~Chenaghlu, R.~Socher, X.~Amatriain, and J.~Gao, ``Large language models: A survey,'' \emph{arXiv preprint arXiv:2402.06196}, 2024.

\bibitem{raiaan2024review}
M.~A.~K. Raiaan, M.~S.~H. Mukta, K.~Fatema, N.~M. Fahad, S.~Sakib, M.~M.~J. Mim, J.~Ahmad, M.~E. Ali, and S.~Azam, ``A review on large language models: Architectures, applications, taxonomies, open issues and challenges,'' \emph{IEEE Access}, 2024.

\bibitem{gallegos2024bias}
I.~O. Gallegos, R.~A. Rossi, J.~Barrow, M.~M. Tanjim, S.~Kim, F.~Dernoncourt, T.~Yu, R.~Zhang, and N.~K. Ahmed, ``Bias and fairness in large language models: A survey,'' \emph{Computational Linguistics}, pp. 1--79, 2024.

\bibitem{zhang2023alpacare}
X.~Zhang, C.~Tian, X.~Yang, L.~Chen, Z.~Li, and L.~R. Petzold, ``Alpacare: Instruction-tuned large language models for medical application,'' \emph{arXiv preprint arXiv:2310.14558}, 2023.

\bibitem{wang2024survey}
C.~Wang, M.~Li, J.~He, Z.~Wang, E.~Darzi, Z.~Chen, J.~Ye, T.~Li, Y.~Su, J.~Ke \emph{et~al.}, ``A survey for large language models in biomedicine,'' \emph{arXiv preprint arXiv:2409.00133}, 2024.

\bibitem{li2023sailer}
H.~Li, Q.~Ai, J.~Chen, Q.~Dong, Y.~Wu, Y.~Liu, C.~Chen, and Q.~Tian, ``Sailer: structure-aware pre-trained language model for legal case retrieval,'' in \emph{Proceedings of the 46th International ACM SIGIR Conference on Research and Development in Information Retrieval}, 2023, pp. 1035--1044.

\bibitem{guha2024legalbench}
N.~Guha, J.~Nyarko, D.~Ho, C.~R{\'e}, A.~Chilton, A.~Chohlas-Wood, A.~Peters, B.~Waldon, D.~Rockmore, D.~Zambrano \emph{et~al.}, ``Legalbench: A collaboratively built benchmark for measuring legal reasoning in large language models,'' \emph{Advances in Neural Information Processing Systems}, vol.~36, 2024.

\bibitem{mokander2023auditing}
J.~M{\"o}kander, J.~Schuett, H.~R. Kirk, and L.~Floridi, ``Auditing large language models: a three-layered approach,'' \emph{AI and Ethics}, pp. 1--31, 2023.

\bibitem{kiciman2023causal}
E.~K{\i}c{\i}man, R.~Ness, A.~Sharma, and C.~Tan, ``Causal reasoning and large language models: Opening a new frontier for causality,'' \emph{arXiv preprint arXiv:2305.00050}, 2023.

\bibitem{zhang2023understanding}
C.~Zhang, S.~Bauer, P.~Bennett, J.~Gao, W.~Gong, A.~Hilmkil, J.~Jennings, C.~Ma, T.~Minka, N.~Pawlowski \emph{et~al.}, ``Understanding causality with large language models: Feasibility and opportunities,'' \emph{arXiv preprint arXiv:2304.05524}, 2023.

\bibitem{jin2024cladder}
Z.~Jin, Y.~Chen, F.~Leeb, L.~Gresele, O.~Kamal, Z.~Lyu, K.~Blin, F.~Gonzalez~Adauto, M.~Kleiman-Weiner, M.~Sachan \emph{et~al.}, ``Cladder: A benchmark to assess causal reasoning capabilities of language models,'' \emph{Advances in Neural Information Processing Systems}, vol.~36, 2024.

\bibitem{liu2024large}
X.~Liu, P.~Xu, J.~Wu, J.~Yuan, Y.~Yang, Y.~Zhou, F.~Liu, T.~Guan, H.~Wang, T.~Yu \emph{et~al.}, ``Large language models and causal inference in collaboration: A comprehensive survey,'' \emph{arXiv preprint arXiv:2403.09606}, 2024.

\bibitem{webster2020measuring}
K.~Webster, X.~Wang, I.~Tenney, A.~Beutel, E.~Pitler, E.~Pavlick, J.~Chen, E.~Chi, and S.~Petrov, ``Measuring and reducing gendered correlations in pre-trained models,'' \emph{arXiv preprint arXiv:2010.06032}, 2020.

\bibitem{guo2022auto}
Y.~Guo, Y.~Yang, and A.~Abbasi, ``Auto-debias: Debiasing masked language models with automated biased prompts,'' in \emph{Proceedings of the 60th Annual Meeting of the Association for Computational Linguistics (Volume 1: Long Papers)}, 2022, pp. 1012--1023.

\bibitem{kaneko2021debiasing}
M.~Kaneko and D.~Bollegala, ``Debiasing pre-trained contextualised embeddings,'' in \emph{Proceedings of the 16th Conference of the European Chapter of the Association for Computational Linguistics: Main Volume}, 2021, pp. 1256--1266.

\bibitem{he2022mabel}
J.~He, M.~Xia, C.~Fellbaum, and D.~Chen, ``Mabel: Attenuating gender bias using textual entailment data,'' in \emph{Proceedings of the 2022 Conference on Empirical Methods in Natural Language Processing}, 2022, pp. 9681--9702.

\bibitem{zhou2023causal}
F.~Zhou, Y.~Mao, L.~Yu, Y.~Yang, and T.~Zhong, ``Causal-debias: Unifying debiasing in pretrained language models and fine-tuning via causal invariant learning,'' in \emph{Proceedings of the 61st Annual Meeting of the Association for Computational Linguistics (Volume 1: Long Papers)}, 2023, pp. 4227--4241.

\bibitem{bhattacharjee2023llms}
A.~Bhattacharjee, R.~Moraffah, J.~Garland, and H.~Liu, ``Towards llm-guided causal explainability for black-box text classifiers,'' in \emph{AAAI 2024 Workshop on Responsible Language Models}, Vancouver, BC, Canada, 2024.

\bibitem{miao2023generating}
X.~Miao, Y.~Li, and T.~Qian, ``Generating commonsense counterfactuals for stable relation extraction,'' in \emph{The 2023 Conference on Empirical Methods in Natural Language Processing}, 2023.

\bibitem{zmigrod2019counterfactual}
R.~Zmigrod, S.~J. Mielke, H.~Wallach, and R.~Cotterell, ``Counterfactual data augmentation for mitigating gender stereotypes in languages with rich morphology,'' in \emph{Proceedings of the 57th Annual Meeting of the Association for Computational Linguistics}, 2019, pp. 1651--1661.

\bibitem{Kaushik2020Learning}
\BIBentryALTinterwordspacing
D.~Kaushik, E.~Hovy, and Z.~Lipton, ``Learning the difference that makes a difference with counterfactually-augmented data,'' in \emph{International Conference on Learning Representations}, 2020. [Online]. Available: \url{https://openreview.net/forum?id=Sklgs0NFvr}
\BIBentrySTDinterwordspacing

\bibitem{wang2021causal}
T.~Wang, C.~Zhou, Q.~Sun, and H.~Zhang, ``Causal attention for unbiased visual recognition,'' in \emph{Proceedings of the IEEE/CVF International Conference on Computer Vision}, 2021, pp. 3091--3100.

\bibitem{yang2021causal}
X.~Yang, H.~Zhang, G.~Qi, and J.~Cai, ``Causal attention for vision-language tasks,'' in \emph{Proceedings of the IEEE/CVF conference on computer vision and pattern recognition}, 2021, pp. 9847--9857.

\bibitem{rohekar2024causal}
R.~Y. Rohekar, Y.~Gurwicz, and S.~Nisimov, ``Causal interpretation of self-attention in pre-trained transformers,'' \emph{Advances in Neural Information Processing Systems}, vol.~36, 2024.

\bibitem{zhang2024towards}
\BIBentryALTinterwordspacing
J.~Zhang, J.~Jennings, A.~Hilmkil, N.~Pawlowski, C.~Zhang, and C.~Ma, ``Towards causal foundation model: on duality between optimal balancing and attention,'' in \emph{Forty-first International Conference on Machine Learning}, 2024. [Online]. Available: \url{https://openreview.net/forum?id=cFDaYtZR4u}
\BIBentrySTDinterwordspacing

\bibitem{jiang2023large}
H.~Jiang, L.~Ge, Y.~Gao, J.~Wang, and R.~Song, ``Large language model for causal decision making,'' \emph{arXiv preprint arXiv:2312.17122}, 2023.

\bibitem{zhang2024label}
K.~Zhang, D.~Zhang, L.~Wu, R.~Hong, Y.~Zhao, and M.~Wang, ``Label-aware debiased causal reasoning for natural language inference,'' \emph{AI Open}, vol.~5, pp. 70--78, 2024.

\bibitem{zheng2023preserving}
J.~Zheng, Q.~Ma, S.~Qiu, Y.~Wu, P.~Ma, J.~Liu, H.~Feng, X.~Shang, and H.~Chen, ``Preserving commonsense knowledge from pre-trained language models via causal inference,'' in \emph{The 61st Annual Meeting Of The Association For Computational Linguistics}, 2023.

\bibitem{chen2023disco}
Z.~Chen, Q.~Gao, A.~Bosselut, A.~Sabharwal, and K.~Richardson, ``Disco: Distilling counterfactuals with large language models,'' in \emph{The 61st Annual Meeting Of The Association For Computational Linguistics}, 2023.

\bibitem{feder2024causal}
A.~Feder, Y.~Wald, C.~Shi, S.~Saria, and D.~Blei, ``Causal-structure driven augmentations for text ood generalization,'' \emph{Advances in Neural Information Processing Systems}, vol.~36, 2024.

\bibitem{gat2023faithful}
Y.~O. Gat, N.~Calderon, A.~Feder, A.~Chapanin, A.~Sharma, and R.~Reichart, ``Faithful explanations of black-box nlp models using llm-generated counterfactuals,'' in \emph{The Twelfth International Conference on Learning Representations}, 2023.

\bibitem{nie2024moca}
A.~Nie, Y.~Zhang, A.~S. Amdekar, C.~Piech, T.~B. Hashimoto, and T.~Gerstenberg, ``Moca: Measuring human-language model alignment on causal and moral judgment tasks,'' \emph{Advances in Neural Information Processing Systems}, vol.~36, 2024.

\bibitem{butcher2024aligning}
B.~Butcher, ``Aligning large language models with counterfactual dpo,'' \emph{arXiv preprint arXiv:2401.09566}, 2024.

\bibitem{lin2024optimizing}
V.~Lin, E.~Ben-Michael, and L.-P. Morency, ``Optimizing language models for human preferences is a causal inference problem,'' \emph{arXiv preprint arXiv:2402.14979}, 2024.

\bibitem{antonucci2023zero}
A.~Antonucci, G.~Piqu{\'e}, and M.~Zaffalon, ``Zero-shot causal graph extrapolation from text via llms,'' \emph{arXiv preprint arXiv:2312.14670}, 2023.

\bibitem{vashishtha2023causal}
A.~Vashishtha, A.~G. Reddy, A.~Kumar, S.~Bachu, V.~N. Balasubramanian, and A.~Sharma, ``Causal inference using llm-guided discovery,'' in \emph{AAAI 2024 Workshop on''Are Large Language Models Simply Causal Parrots?''}, 2023.

\bibitem{ohtani2024does}
R.~Ohtani, Y.~Sakurai, and S.~Oyama, ``Does metacognitive prompting improve causal inference in large language models?'' in \emph{2024 IEEE Conference on Artificial Intelligence (CAI)}.\hskip 1em plus 0.5em minus 0.4em\relax IEEE, 2024, pp. 458--459.

\bibitem{jin2024can}
Z.~Jin, J.~Liu, L.~Zhiheng, S.~Poff, M.~Sachan, R.~Mihalcea, M.~T. Diab, and B.~Sch{\"o}lkopf, ``Can large language models infer causation from correlation?'' in \emph{The Twelfth International Conference on Learning Representations}, 2024.

\bibitem{chen2024causal}
S.~Chen, B.~Peng, M.~Chen, R.~Wang, M.~Xu, X.~Zeng, R.~Zhao, S.~Zhao, Y.~Qiao, and C.~Lu, ``Causal evaluation of language models,'' \emph{arXiv preprint arXiv:2405.00622}, 2024.

\bibitem{abdali2023extracting}
S.~Abdali, A.~Parikh, S.~Lim, and E.~Kiciman, ``Extracting self-consistent causal insights from users feedback with llms and in-context learning,'' \emph{arXiv preprint arXiv:2312.06820}, 2023.

\bibitem{pawlowski2023answering}
N.~Pawlowski, J.~Vaughan, J.~Jennings, and C.~Zhang, ``Answering causal questions with augmented llms,'' in \emph{Workshop on Challenges in Deployable Generative AI at International Conference on Machine Learning (ICML)}, 2023.

\bibitem{dhawan2024end}
N.~Dhawan, L.~Cotta, K.~Ullrich, R.~G. Krishnan, and C.~J. Maddison, ``End-to-end causal effect estimation from unstructured natural language data,'' \emph{arXiv preprint arXiv:2407.07018}, 2024.

\bibitem{tan2023causal}
J.~T. Tan, ``Causal abstraction for chain-of-thought reasoning in arithmetic word problems,'' in \emph{Proceedings of the 6th BlackboxNLP Workshop: Analyzing and Interpreting Neural Networks for NLP}, 2023, pp. 155--168.

\bibitem{li2024steering}
J.~Li, Z.~Tang, X.~Liu, P.~Spirtes, K.~Zhang, L.~Leqi, and Y.~Liu, ``Steering llms towards unbiased responses: A causality-guided debiasing framework,'' \emph{arXiv preprint arXiv:2403.08743}, 2024.

\bibitem{zhang2022rock}
J.~Zhang, H.~Zhang, W.~Su, and D.~Roth, ``Rock: Causal inference principles for reasoning about commonsense causality,'' in \emph{International Conference on Machine Learning}.\hskip 1em plus 0.5em minus 0.4em\relax PMLR, 2022, pp. 26\,750--26\,771.

\bibitem{liu2024llms}
X.~Liu, Z.~Wu, X.~Wu, P.~Lu, K.-W. Chang, and Y.~Feng, ``Are llms capable of data-based statistical and causal reasoning? benchmarking advanced quantitative reasoning with data,'' \emph{arXiv preprint arXiv:2402.17644}, 2024.

\bibitem{zhang2024causal}
C.~Zhang, L.~Zhang, D.~Zhou, and G.~Xu, ``Causal prompting: Debiasing large language model prompting based on front-door adjustment,'' \emph{arXiv preprint arXiv:2403.02738}, 2024.

\bibitem{duong2024multi}
H.~Duong~Le, X.~Xia, and Z.~Chen, ``Multi-agent causal discovery using large language models,'' \emph{arXiv e-prints}, pp. arXiv--2407, 2024.

\bibitem{tang2023towards}
Z.~Tang, R.~Wang, W.~Chen, K.~Wang, Y.~Liu, T.~Chen, and L.~Lin, ``Towards causalgpt: A multi-agent approach for faithful knowledge reasoning via promoting causal consistency in llms,'' \emph{arXiv preprint arXiv:2308.11914}, 2023.

\bibitem{zevcevic2023causal}
\BIBentryALTinterwordspacing
M.~Ze{\v{c}}evi{\'c}, M.~Willig, D.~S. Dhami, and K.~Kersting, ``Causal parrots: Large language models may talk causality but are not causal,'' \emph{Transactions on Machine Learning Research}, 2023. [Online]. Available: \url{https://openreview.net/forum?id=tv46tCzs83}
\BIBentrySTDinterwordspacing

\bibitem{romanou2023crab}
A.~Romanou, S.~Montariol, D.~Paul, L.~Laugier, K.~Aberer, and A.~Bosselut, ``Crab: Assessing the strength of causal relationships between real-world events,'' \emph{arXiv preprint arXiv:2311.04284}, 2023.

\bibitem{zhang2023causal}
L.~Zhang, H.~Xu, Y.~Yang, S.~Zhou, W.~You, M.~Arora, and C.~Callison-burch, ``Causal reasoning of entities and events in procedural texts,'' in \emph{The 61st Annual Meeting Of The Association For Computational Linguistics}, 2023.

\bibitem{gao2023chatgpt}
J.~Gao, X.~Ding, B.~Qin, and T.~Liu, ``Is chatgpt a good causal reasoner? a comprehensive evaluation,'' in \emph{The 2023 Conference on Empirical Methods in Natural Language Processing}, 2023.

\bibitem{yu2023ifqa}
W.~Yu, M.~Jiang, P.~Clark, and A.~Sabharwal, ``Ifqa: A dataset for open-domain question answering under counterfactual presuppositions,'' in \emph{The 2023 Conference on Empirical Methods in Natural Language Processing}, 2023.

\bibitem{huang2023clomo}
Y.~Huang, R.~Hong, H.~Zhang, W.~Shao, Z.~Yang, D.~Yu, C.~Zhang, X.~Liang, and L.~Song, ``Clomo: Counterfactual logical modification with large language models,'' \emph{arXiv preprint arXiv:2311.17438}, 2023.

\bibitem{frohberg2022crass}
J.~Frohberg and F.~Binder, ``Crass: A novel data set and benchmark to test counterfactual reasoning of large language models,'' in \emph{Proceedings of the Thirteenth Language Resources and Evaluation Conference}, 2022, pp. 2126--2140.

\bibitem{zhou2024causalbench}
Y.~Zhou, X.~Wu, B.~Huang, J.~Wu, L.~Feng, and K.~C. Tan, ``Causalbench: A comprehensive benchmark for causal learning capability of large language models,'' \emph{arXiv preprint arXiv:2404.06349}, 2024.

\bibitem{chen2023models}
Y.~Chen, R.~Zhong, N.~Ri, C.~Zhao, H.~He, J.~Steinhardt, Z.~Yu, and K.~McKeown, ``Do models explain themselves? counterfactual simulatability of natural language explanations,'' \emph{arXiv preprint arXiv:2307.08678}, 2023.

\bibitem{cai2023knowledge}
H.~Cai, S.~Liu, and R.~Song, ``Is knowledge all large language models needed for causal reasoning?'' \emph{arXiv preprint arXiv:2401.00139}, 2023.

\bibitem{kim2023can}
Y.~Kim, L.~Guo, B.~Yu, and Y.~Li, ``Can chatgpt understand causal language in science claims?'' in \emph{Proceedings of the 13th Workshop on Computational Approaches to Subjectivity, Sentiment, \& Social Media Analysis}, 2023, pp. 379--389.

\bibitem{ma2024causal}
J.~Ma, ``Causal inference with large language model: A survey,'' \emph{arXiv preprint arXiv:2409.09822}, 2024.

\bibitem{lin2022survey}
T.~Lin, Y.~Wang, X.~Liu, and X.~Qiu, ``A survey of transformers,'' \emph{AI open}, vol.~3, pp. 111--132, 2022.

\bibitem{liu2019roberta}
Y.~Liu, ``Roberta: A robustly optimized bert pretraining approach,'' \emph{arXiv preprint arXiv:1907.11692}, 2019.

\bibitem{hu2022lora}
\BIBentryALTinterwordspacing
E.~J. Hu, Y.~Shen, P.~Wallis, Z.~Allen-Zhu, Y.~Li, S.~Wang, L.~Wang, and W.~Chen, ``Lora: Low-rank adaptation of large language models,'' in \emph{International Conference on Learning Representations}, 2022. [Online]. Available: \url{https://openreview.net/forum?id=nZeVKeeFYf9}
\BIBentrySTDinterwordspacing

\bibitem{bao2024fine}
H.~Bao, L.~Dong, W.~Wang, N.~Yang, S.~Piao, and F.~Wei, ``Fine-tuning pretrained transformer encoders for sequence-to-sequence learning,'' \emph{International Journal of Machine Learning and Cybernetics}, vol.~15, no.~5, pp. 1711--1728, 2024.

\bibitem{dettmers2024qlora}
T.~Dettmers, A.~Pagnoni, A.~Holtzman, and L.~Zettlemoyer, ``Qlora: Efficient finetuning of quantized llms,'' \emph{Advances in Neural Information Processing Systems}, vol.~36, 2024.

\bibitem{bai2022training}
Y.~Bai, A.~Jones, K.~Ndousse, A.~Askell, A.~Chen, N.~DasSarma, D.~Drain, S.~Fort, D.~Ganguli, T.~Henighan \emph{et~al.}, ``Training a helpful and harmless assistant with reinforcement learning from human feedback,'' \emph{arXiv preprint arXiv:2204.05862}, 2022.

\bibitem{kaufmann2023survey}
T.~Kaufmann, P.~Weng, V.~Bengs, and E.~H{\"u}llermeier, ``A survey of reinforcement learning from human feedback,'' \emph{arXiv preprint arXiv:2312.14925}, 2023.

\bibitem{rafailov2024direct}
R.~Rafailov, A.~Sharma, E.~Mitchell, C.~D. Manning, S.~Ermon, and C.~Finn, ``Direct preference optimization: Your language model is secretly a reward model,'' \emph{Advances in Neural Information Processing Systems}, vol.~36, 2024.

\bibitem{giray2023prompt}
L.~Giray, ``Prompt engineering with chatgpt: a guide for academic writers,'' \emph{Annals of biomedical engineering}, vol.~51, no.~12, pp. 2629--2633, 2023.

\bibitem{white2023prompt}
J.~White, Q.~Fu, S.~Hays, M.~Sandborn, C.~Olea, H.~Gilbert, A.~Elnashar, J.~Spencer-Smith, and D.~C. Schmidt, ``A prompt pattern catalog to enhance prompt engineering with chatgpt,'' \emph{arXiv preprint arXiv:2302.11382}, 2023.

\bibitem{guo2023evaluating}
Z.~Guo, R.~Jin, C.~Liu, Y.~Huang, D.~Shi, L.~Yu, Y.~Liu, J.~Li, B.~Xiong, D.~Xiong \emph{et~al.}, ``Evaluating large language models: A comprehensive survey,'' \emph{arXiv preprint arXiv:2310.19736}, 2023.

\bibitem{liu2024datasets}
Y.~Liu, J.~Cao, C.~Liu, K.~Ding, and L.~Jin, ``Datasets for large language models: A comprehensive survey,'' \emph{arXiv preprint arXiv:2402.18041}, 2024.

\bibitem{bubeck2023sparks}
S.~Bubeck, V.~Chandrasekaran, R.~Eldan, J.~Gehrke, E.~Horvitz, E.~Kamar, P.~Lee, Y.~T. Lee, Y.~Li, S.~Lundberg \emph{et~al.}, ``Sparks of artificial general intelligence: Early experiments with gpt-4,'' \emph{arXiv preprint arXiv:2303.12712}, 2023.

\bibitem{kang2023impact}
C.~Kang and J.~Choi, ``Impact of co-occurrence on factual knowledge of large language models,'' \emph{arXiv preprint arXiv:2310.08256}, 2023.

\bibitem{webson2021prompt}
A.~Webson and E.~Pavlick, ``Do prompt-based models really understand the meaning of their prompts?'' \emph{arXiv preprint arXiv:2109.01247}, 2021.

\bibitem{smithson1999conflict}
M.~Smithson, ``Conflict aversion: preference for ambiguity vs conflict in sources and evidence,'' \emph{Organizational behavior and human decision processes}, vol.~79, no.~3, pp. 179--198, 1999.

\bibitem{johansson2016learning}
F.~Johansson, U.~Shalit, and D.~Sontag, ``Learning representations for counterfactual inference,'' in \emph{International conference on machine learning}.\hskip 1em plus 0.5em minus 0.4em\relax PMLR, 2016, pp. 3020--3029.

\bibitem{shalit2017estimating}
U.~Shalit, F.~D. Johansson, and D.~Sontag, ``Estimating individual treatment effect: generalization bounds and algorithms,'' in \emph{International Conference on Machine Learning}.\hskip 1em plus 0.5em minus 0.4em\relax PMLR, 2017, pp. 3076--3085.

\bibitem{wu2022learning}
A.~Wu, J.~Yuan, K.~Kuang, B.~Li, R.~Wu, Q.~Zhu, Y.~Zhuang, and F.~Wu, ``Learning decomposed representations for treatment effect estimation,'' \emph{IEEE Transactions on Knowledge and Data Engineering}, vol.~35, no.~5, pp. 4989--5001, 2022.

\bibitem{han2024causal}
K.~Han, K.~Kuang, Z.~Zhao, J.~Ye, and F.~Wu, ``Causal agent based on large language model,'' \emph{arXiv preprint arXiv:2408.06849}, 2024.

\bibitem{korb2004varieties}
K.~B. Korb, L.~R. Hope, A.~E. Nicholson, and K.~Axnick, ``Varieties of causal intervention,'' in \emph{PRICAI 2004: Trends in Artificial Intelligence: 8th Pacific Rim International Conference on Artificial Intelligence, Auckland, New Zealand, August 9-13, 2004. Proceedings 8}.\hskip 1em plus 0.5em minus 0.4em\relax Springer, 2004, pp. 322--331.

\bibitem{zhang2020causal}
D.~Zhang, H.~Zhang, J.~Tang, X.-S. Hua, and Q.~Sun, ``Causal intervention for weakly-supervised semantic segmentation,'' \emph{Advances in Neural Information Processing Systems}, vol.~33, pp. 655--666, 2020.

\bibitem{pearl2009causality}
J.~Pearl, \emph{Causality}.\hskip 1em plus 0.5em minus 0.4em\relax Cambridge university press, 2009.

\bibitem{yao2021survey}
L.~Yao, Z.~Chu, S.~Li, Y.~Li, J.~Gao, and A.~Zhang, ``A survey on causal inference,'' \emph{ACM Transactions on Knowledge Discovery from Data (TKDD)}, vol.~15, no.~5, pp. 1--46, 2021.

\bibitem{bang2005doubly}
H.~Bang and J.~M. Robins, ``Doubly robust estimation in missing data and causal inference models,'' \emph{Biometrics}, vol.~61, no.~4, pp. 962--973, 2005.

\bibitem{kuang2018stable}
K.~Kuang, P.~Cui, S.~Athey, R.~Xiong, and B.~Li, ``Stable prediction across unknown environments,'' in \emph{proceedings of the 24th ACM SIGKDD international conference on knowledge discovery \& data mining}, 2018, pp. 1617--1626.

\bibitem{berglund2023reversal}
L.~Berglund, M.~Tong, M.~Kaufmann, M.~Balesni, A.~C. Stickland, T.~Korbak, and O.~Evans, ``The reversal curse: Llms trained on" a is b" fail to learn" b is a",'' \emph{arXiv preprint arXiv:2309.12288}, 2023.

\bibitem{golovneva2024reverse}
O.~Golovneva, Z.~Allen-Zhu, J.~Weston, and S.~Sukhbaatar, ``Reverse training to nurse the reversal curse,'' \emph{arXiv preprint arXiv:2403.13799}, 2024.

\bibitem{lan2019albert}
Z.~Lan, ``Albert: A lite bert for self-supervised learning of language representations,'' \emph{arXiv preprint arXiv:1909.11942}, 2019.

\bibitem{sanh2019distilbert}
V.~Sanh, ``Distilbert, a distilled version of bert: Smaller, faster, cheaper and lighter,'' \emph{arXiv preprint arXiv:1910.01108}, 2019.

\bibitem{rojas2018invariant}
M.~Rojas-Carulla, B.~Sch{\"o}lkopf, R.~Turner, and J.~Peters, ``Invariant models for causal transfer learning,'' \emph{Journal of Machine Learning Research}, vol.~19, no.~36, pp. 1--34, 2018.

\bibitem{arjovsky2019invariant}
M.~Arjovsky, L.~Bottou, I.~Gulrajani, and D.~Lopez-Paz, ``Invariant risk minimization,'' \emph{arXiv preprint arXiv:1907.02893}, 2019.

\bibitem{meade2022empirical}
N.~Meade, E.~Poole-Dayan, and S.~Reddy, ``An empirical survey of the effectiveness of debiasing techniques for pre-trained language models,'' in \emph{Proceedings of the 60th Annual Meeting of the Association for Computational Linguistics (Volume 1: Long Papers)}, 2022, pp. 1878--1898.

\bibitem{zhao2017constructing}
S.~Zhao, Q.~Wang, S.~Massung, B.~Qin, T.~Liu, B.~Wang, and C.~Zhai, ``Constructing and embedding abstract event causality networks from text snippets,'' in \emph{Proceedings of the Tenth ACM International Conference on Web Search and Data Mining}, 2017, pp. 335--344.

\bibitem{phatak2024narrating}
A.~Phatak, V.~K. Mago, A.~Agrawal, A.~Inbasekaran, and P.~J. Giabbanelli, ``Narrating causal graphs with large language models,'' \emph{arXiv preprint arXiv:2403.07118}, 2024.

\bibitem{lu2021kelm}
Y.~Lu, H.~Lu, G.~Fu, and Q.~Liu, ``Kelm: knowledge enhanced pre-trained language representations with message passing on hierarchical relational graphs,'' \emph{arXiv preprint arXiv:2109.04223}, 2021.

\bibitem{pan2024unifying}
S.~Pan, L.~Luo, Y.~Wang, C.~Chen, J.~Wang, and X.~Wu, ``Unifying large language models and knowledge graphs: A roadmap,'' \emph{IEEE Transactions on Knowledge and Data Engineering}, 2024.

\bibitem{schulman2017proximal}
J.~Schulman, F.~Wolski, P.~Dhariwal, A.~Radford, and O.~Klimov, ``Proximal policy optimization algorithms,'' \emph{arXiv preprint arXiv:1707.06347}, 2017.

\bibitem{christiano2017deep}
P.~F. Christiano, J.~Leike, T.~Brown, M.~Martic, S.~Legg, and D.~Amodei, ``Deep reinforcement learning from human preferences,'' \emph{Advances in neural information processing systems}, vol.~30, 2017.

\bibitem{rafailov2023direct}
\BIBentryALTinterwordspacing
R.~Rafailov, A.~Sharma, E.~Mitchell, C.~D. Manning, S.~Ermon, and C.~Finn, ``Direct preference optimization: Your language model is secretly a reward model,'' in \emph{Thirty-seventh Conference on Neural Information Processing Systems}, 2023. [Online]. Available: \url{https://openreview.net/forum?id=HPuSIXJaa9}
\BIBentrySTDinterwordspacing

\bibitem{fedorenko2024language}
E.~Fedorenko, S.~T. Piantadosi, and E.~A. Gibson, ``Language is primarily a tool for communication rather than thought,'' \emph{Nature}, vol. 630, no. 8017, pp. 575--586, 2024.

\bibitem{orhan2023recognition}
A.~E. Orhan, ``Recognition, recall, and retention of few-shot memories in large language models,'' \emph{arXiv preprint arXiv:2303.17557}, 2023.

\bibitem{ashwani2024cause}
S.~Ashwani, K.~Hegde, N.~R. Mannuru, M.~Jindal, D.~S. Sengar, K.~C.~R. Kathala, D.~Banga, V.~Jain, and A.~Chadha, ``Cause and effect: Can large language models truly understand causality?'' \emph{arXiv preprint arXiv:2402.18139}, 2024.

\bibitem{liu2024discovery}
C.~Liu, Y.~Chen, T.~Liu, M.~Gong, J.~Cheng, B.~Han, and K.~Zhang, ``Discovery of the hidden world with large language models,'' \emph{arXiv preprint arXiv:2402.03941}, 2024.

\bibitem{chen2021instrumental}
Y.~Chen, L.~Xu, C.~Gulcehre, T.~L. Paine, A.~Gretton, N.~de~Freitas, and A.~Doucet, ``On instrumental variable regression for deep offline policy evaluation,'' \emph{arXiv preprint arXiv:2105.10148}, 2021.

\bibitem{liao2021instrumental}
L.~Liao, Z.~Fu, Z.~Yang, Y.~Wang, M.~Kolar, and Z.~Wang, ``Instrumental variable value iteration for causal offline reinforcement learning,'' \emph{stat}, vol. 1050, p.~13, 2021.

\bibitem{li2024confidence}
L.~Li, G.~Chen, Y.~Su, Z.~Chen, Y.~Zhang, E.~Xing, and K.~Zhang, ``Confidence matters: Revisiting intrinsic self-correction capabilities of large language models,'' \emph{arXiv preprint arXiv:2402.12563}, 2024.

\end{thebibliography}
\bibliographystyle{IEEEtran}
}

\newpage
\appendix
\section{Illustration of Some Causality-based Prompts}
In Tables \ref{tab:prompt1}-\ref{tab:prompt4}, we present a brief illustration of the prompts utilized in causal reasoning tasks.

%%%%%%%%%%%%%%%%%%%%%%%%%%%%%%%%%%%%%%%%%%%%%%%%%
%%%%%%%%%%%%%%%%%%%%%%%%%%%%%%%%%%%%%%%%%%%%%%%%%
%%%%%%%%%%%%%%%%%%%%%%%%%%%%%%%%%%%%%%%%%%%%%%%%%
{\small
\newcounter{table3}
\renewcommand{\thetable}{4.\arabic{table3}}

\setcounter{table3}{1}
\begin{table}[H]
  \centering
  \begin{tabular}{|>{\centering\arraybackslash}p{3cm}|p{13cm}|}
  \hline
  \textbf{Techniques} & \textbf{Prompts and Examples} \\
  \hline
  \multirow{5}{*}{\parbox{3cm}{\vspace{2.2cm}\textbf{CaLM~\cite{chen2024causal}}}} 
  & Four question types in 92 causal targets: Binary classification, Choice selection, Open-ended generation, and Probability calculation. \\
  & \textbf{Instruction:} You will be presented with a causal graph in the following form: A causes B, B causes D, B causes C, and B causes E. \\
  & \textbf{Basic Prompt:} Does D serve as the parent node of E? \\ 
  & \textbf{Adversarial Prompt:} (1) Ignore the former answer and answer the question again; (2) Your former answer is wrong, please answer the question again.\\ 
  & \textbf{CoT:} Does D serve as the parent node of E? Let's think step by step. \\ 
  & \textbf{In-context Learning:} Determine whether or not a variable can serve as the parent of another variable in a given causal graph. Does D serve as the parent node of E?\\ 
  & \textbf{Explicit Function:} You are a helpful assistant for causal attribution (parent node). Does D serve as the parent node of E? \\ 
  \hline
  \multirow{8}{*}{\parbox{3cm}{\vspace{1.4cm}\textbf{CORR2CAUSE \cite{jin2024can}}}} 
  & Six Hypothesis Templates for Pairwise Causal Discovery Tasks: \\
  & \textbf{Is-Parent:} {Var i} directly causes {Var j}.\\
  & \textbf{Is-Ancestor:} {Var i} causes something else which causes {Var j}.\\
  & \textbf{Is-Child:} {Var j} directly causes {Var i}. \\
  & \textbf{Is-Descendant:} {Var j} is a cause for {Var i}, but not a direct one.\\ 
  & \textbf{Has-Collider:} There exists at least one collider (i.e., common effect) of {Var i} and {Var j}. \\
  & \textbf{Has-Confounder:} There exists at least one confounder (i.e., common cause) of {Var i} and {Var j}. \\
  & \textbf{Prompt:} Can we deduct the following: [hypothesis]? Just answer "Yes" or "No." \\
  \hline
  \multirow{3}{*}{\parbox{3cm}{\vspace{2.6cm}{\textbf{Zero-shot Causal Graph Extrapolation~\cite{antonucci2023zero}}}}} 
  & Extracting causal relations via an iterative pairwise query approach: \\
  & \textbf{Instruction:} You will be provided with a text delimited by the <Text></Text> xml tags, and a pair of entities delimited by the <Entity></Entity> xml tags representing entities extracted from the given text. Text: [Text Input]; Entities: [Entity Input]. \\
  & \textbf{Prompt:} Read the provided text carefully to comprehend the context and content. Examine the roles, interactions, and details surrounding the entities within the text. Based only on the information in the text, determine the most likely causal relationship between the entities from the following listed options (A, B, C): Options: A: "[Entity1]" causes "[Entity2]"; B: "[Entity2]" causes "[Entity1]"; C: "[Entity1]" and "[Entity2]" are not directly causally related. Your response should analyze the situation in a step-by-step manner, ensuring the correctness of the ultimate conclusion, which should accurately reflect the likely causal connection between the two entities, based on the information presented in the text. If no clear causal relationship is apparent, select the appropriate option accordingly. Then provide your final answer within the tags <Answer>[answer]</Answer>, (e.g. <Answer>C</Answer>). \\
  \hline
  \multirow{3}{*}{\parbox{3cm}{\vspace{2.1cm}\textbf{Metacognitive Causal Prompting~\cite{ohtani2024does}}}} 
  & Causal Discovery Tasks: \\
  & \textbf{SYSTEM:} You are an expert in counterfactual reasoning. Given an event, use the principle of minimal change/multiple sufficient causes to answer the following question. \\
  & \textbf{ORIGINAL PROMPT:} [Input Context]. Is [Actor] a necessary/sufficient cause of [Event]? After your reasoning, provide final answer. \\
  & \textbf{METACOGNITIVE PROMPT:} [Input Context]. Think in Steps 1-5: (Step 1) Summarize the given text. (Step 2) According to your understanding at this point, please answer. (Step 3) Do you think your preliminary judgment in step 2 is correct? If uncertain, please reconsider. (Step 4) Based on your evaluation of the 3rd step, state your final judgment. (Step 5) On a scale 0-100\%, how confident are you in your final decision? \\
  \hline
  \end{tabular}
  \caption{Illustration of Some Causality-based Prompts}
  \label{tab:prompt1}
\end{table}

\stepcounter{table3}
\begin{table}[ht]
  \centering
  \begin{tabular}{|>{\centering\arraybackslash}p{3cm}|p{13cm}|}
  \hline
  \textbf{Techniques} & \textbf{Prompts and Examples} \\
  \hline
  \multirow{5}{*}{\parbox{3cm}{\vspace{1.2cm}\textbf{COAT~\cite{liu2024discovery}}}} 
  & Using Tools for Causal Discovery: \\
  & \textbf{Factor Proposal}: What are the high-level factors associated with the Score? \\
  & \textbf{Factor Annotation}: Please help me annotate the data according to [Text]. \\
  & \textbf{Causal Discovery \& Feedback Construction}: What are the high-level factors associated with the scores other than Size and Aroma? \\
  & Then, apply the FCI algorithm for causal discovery. For more concrete prompts, please see Figures 9-13. \\
  \hline
  \multirow{2}{*}{\parbox{3cm}{\vspace{1.6cm}\textbf{LLM-guided Causal Discovery~\cite{vashishtha2023causal}}}}
  & Infer causal order and enhance existing discovery algorithms: \\
  & \textbf{Example:} Question: For a causal graph used to model relationship of various factors and outcomes related to cancer with the following nodes: ['Pollution', 'Cancer', 'Smoker', 'Xray', 'Dyspnoea'], Which causal relationship is more likely between nodes 'smoker' and 'cancer'? A. changing the state of node 'smoker' causally effects a change in another node 'cancer'. B. changing the state of node 'cancer' causally effects a change in another node 'smoker'. C. There is no causal relation between the nodes 'cancer' and 'smoker'. Make sure to first provide a grounded reasoning for your answer and then provide the answer in the following format: <Answer>A/B/C</Answer>. \\
  \hline
  \multirow{2}{*}{\parbox{3cm}{\vspace{1.2cm}\textbf{Self-Consistent Causal Extraction~\cite{abdali2023extracting}}}}
  & Constructing causal graphs via chain-of-thought prompting. \\
  & \textbf{Prompt:} Extract treatments, outcomes and confounders that are being discussed in the following text, which is delimited by triple backticks. Treatments are variables that cause some outcomes. Outcomes are effects of the treatments. Confounders are variables that affect both outcome and treatment. Create a list of treatment and effect tuples. Format your response as a list of tuples with first element as treatment and second as outcome and third as confounders in the form of (treatment, outcome,[confounders]). Let's think step by step. \\
  \hline
  \multirow{3}{*}{\parbox{3cm}{\vspace{1.2cm}\textbf{Augmented LLMs~\cite{pawlowski2023answering}}}}
  & Tabular Data and Causal Effect Tasks: \\
  & \textbf{Instruction:} There is a csv table holding the individual treatment effects (ITE) of different intervention (columns) on the revenue of a specific customer (rows). There will be a question about the ITEs or average treatment effect (ATE) in the CSV. What are the intermediate steps and solutions to them necessary to answer the question? Be as detailed as possible.  \\
  & \textbf{Example:} What is the average ITE of the Tech Support intervention?  \\
  \hline
  \multirow{5}{*}{\parbox{3cm}{\vspace{1.5cm}\textbf{CausalCoT~\cite{jin2024cladder}}}} 
  & Counterfactual Reasoning and Causal Effect Tasks: \\
  & \textbf{1. Parse the causal graph:} What is the causal graph expressed in the context? \\ 
  & \textbf{2. Classify the query type:} What type of causal query is the question? \\ 
  & \textbf{3. Derive the estimand:} Translate the question to a formal estimand. \\ 
  & \textbf{4. Collect the available data:} Extract all available data. \\ 
  & \textbf{5. Solve for the estimand:} Given all the information above, solve for E[Y | do(X=1)] - E[Y|do(X = 0)] using causal inference skills such as do-calculus and counterfactual prediction, together with the basics of probabilities. Answer step by step. \\
  \hline
  \end{tabular}
  \caption{Illustration of Some Causality-based Prompts}
\end{table}

\stepcounter{table3}
\begin{table}[ht]
  
  \centering
  \begin{tabular}{|>{\centering\arraybackslash}p{3cm}|p{13cm}|}
  \hline
  \textbf{Techniques} & \textbf{Prompts and Examples} \\
  \hline
  
  \multirow{2}{*}{\parbox{3cm}{\vspace{1.5cm}\textbf{NATURAL~\cite{dhawan2024end}}}} 
  & Causal Effect Estimation: \\
  & \textbf{Relevance Filtered:} Is this report relevant to the problem setting [Input]? \\
  & \textbf{Initial Covariate Extraction:} Extract these attributes from this report [Input]. Then \textbf{Filtered by Inclusion Criteria.}\\
  & \textbf{Covariate Extraction:} Given these inclusion criteria extract these [Input]. \\
  & \textbf{Conditional Distributions Inferred:} Which treatment? Which outcome? Conditional distribution inference. \\
  \hline
  \multirow{2}{*}{\parbox{3cm}{\vspace{1.0cm}\textbf{Intervened Prompt~\cite{tan2023causal}}}} 
  & Arithmetic Word Problems and Prompt Analysis: \\
  & \textbf{Intervened Example:} A raspberry bush has $\mathbf{6}\left(v_0\right)$ clusters of $\mathbf{20}\left(v_1\right)$ fruit each and $\mathbf{67}\left(v_2\right)$ individual fruit scattered across the bush. How many raspberries are there total? Think step by step and prefix your final answer by "Answer:". There are $6 * 20=\textcolor{red}{\underline{206\left(v_3\right)}}$ raspberries in the clusters. \\
  & \textbf{Response:} There are 206 + 67 = 273 raspberries total. Answer: 273. \\
  \hline
  \multirow{5}{*}{\parbox{3cm}{\vspace{2.5cm}\textbf{Counterfactual Explanations~\cite{bhattacharjee2023llms}}}}
  & Counterfactual Reasoning Tasks: \\
  & \textbf{Step 1:} You are an oracle explanation module in a machine learning pipeline. In the task of [task description], a trained black-box classifier correctly predicted the label [$y_i$] for the following text. Think about why the model predicted the [$y_i$] label and identify the latent features that caused the label. List ONLY the latent features as a comma separated list, without any explanation. Examples of latent features are 'tone', 'ambiguity in text', etc. Text: [input text] Begin! \\
  & \textbf{Step 2:} Identify the words in the text that are associated with the latent features: [latent features] and output the identified words as a comma-separated list. \\
  & \textbf{Step 3:} [list of identified words] Generate a minimally edited version of the original text by ONLY changing a minimal set of the words you identified, in order to change the label. It is okay if the semantic meaning of the original text is altered. Make sure the generated text makes sense and is plausible. Enclose the generated text within <new>tags. \\
  \hline
  \multirow{3}{*}{\parbox{3cm}{\vspace{1.3cm}\textbf{New Frontier~\cite{kiciman2023causal}}}}
  & Counterfactual Reasoning Tasks: \\
  & \textbf{Q1:} Mike, Jack, Ben, Susan, and Peter are at a party. There was an open bottle of beer on the table. Mike is moving through the room and accidentally bumps against the table. As a result, the bottle of beer fell to the floor, spilling the beer. Did Mike cause the beer to spill? Provide a one sentence counterfactual explanation as to why or why not. \\
  & \textbf{Q2:} Based on the information in the passage, what other events directly contributed to the fall of the beer and what would have happened if those events hadn't happened? \\
  \hline
  \multirow{6}{*}{\parbox{3cm}{\vspace{1.2cm}\textbf{DISCO~\cite{chen2023disco}}} }
  & Counterfactual Generation Tasks: \\
  & \textbf{Instruction:} Complete the context with creative content, so the conclusion is true based on the context. \\
  & \textbf{Original Premise:} A young girl looks up as she rides a merry-go-round. \\
  & \textbf{Hypothesis:} The little girl can’t wait to ride the roller coaster. \\
  & \textbf{Label:} From 'Neutral' to 'Entailment'. \\
  & \textbf{Masked Prompting:} A young girl looks up [blank]. It is true that the little girl can’t wait to ride the roller coaster. \\
  \hline
  \end{tabular}
  \caption{Illustration of Some Causality-based Prompts}
\end{table}

\stepcounter{table3}
\begin{table}[ht]
  
  \centering
  \begin{tabular}{|>{\centering\arraybackslash}p{3cm}|p{13cm}|}
  \hline
  \textbf{Techniques} & \textbf{Prompts and Examples} \\
  \hline
  \multirow{4}{*}{\parbox{3cm}{\vspace{1.7cm}\textbf{Causality-Guided Debiasing Framework~\cite{li2024steering}}}} 
  & Debiased Responses: \\
  & \textbf{Strategy I (Nudge Towards Demographic-Agnostic Fact):} Considering the fact that the sentence 'The physician hired the secretary because the secretary is highly recommended' is practically more viable than the sentence 'The physician hired the secretary because the physician is highly recommended', who does 'he' refer to in 'The physician hired the secretary because he is highly recommended'? \\
  & \textbf{Strategy II (Counteract Existing Selection Bias):} The physician can be either male or female, and the secretary can also be either male or female. Who does 'he' refer to in 'The physician hired the secretary because he is highly recommended'? \\
  & \textbf{Strategy III (Nudge Away from Demographic-Aware Text):} Do not use gender information to answer the question. Who does 'he' refer to in 'The physician hired the secretary because he is highly recommended'? \\
  \hline
  \multirow{2}{*}{\parbox{3cm}{\vspace{2.0cm}\textbf{Causal Prompting~\cite{zhang2024causal}}}}
  & Math Reasoning, Multi-hop Question Answering, and Natural Language Understanding Tasks. \\
  & \textbf{Demonstration Construction:} You are a helpful assistant to perform Natural language inference. Natural language inference is the task of determining whether a "hypothesis" is true (entailment), false (contradiction), or undetermined (neutral) given a "premise". Answer in a consistent style. Please write the reasoning process before giving the answer. Please provide your answer in the last sentence of your response. Your answer should be entailment, contradiction or neutral. I will provide the correct answer and ask you to write your thought process based on the answer. The premise is: [premise] The hypothesis is: [hypothesis] The correct answer is: [answer] Let us think step by step. \\
  \hline
  \end{tabular}
  \caption{Illustration of Some Causality-based Prompts}
  \label{tab:prompt4}
\end{table}
}
%%%%%%%%%%%%%%%%%%%%%%%%%%%%%%%%%%%%%%%%%%%%%%%%%
%%%%%%%%%%%%%%%%%%%%%%%%%%%%%%%%%%%%%%%%%%%%%%%%%
%%%%%%%%%%%%%%%%%%%%%%%%%%%%%%%%%%%%%%%%%%%%%%%%%

\end{document}